%% file: main.tex
\pdfoutput=1
\documentclass[acmsmall,screen,nonacm]{acmart}
\AtBeginDocument{%
  }

\acmJournal{TACO}
\acmVolume{0}
\acmNumber{0}
\acmArticle{0}
\acmMonth{0}
\acmYear{0}

\input{contents/environment.tex}

\usepackage{booktabs}

\begin{document}

\title{From Overloaded to Guaranteed: High-Throughput Multi-SLO Enforcement for LoRA-Assisted On-Premise LLM Deployment}

\author{Zeshen Zhang}
\email{gdeepolar@gmail.com}
\author{Han Zhao}
\authornote{Corresponding author.}
\email{zhao-han@cs.sjtu.edu.cn}
\author{Weihao Cui}
\email{weihao@sjtu.edu.cn}
\author{Quan Chen}
\email{chen-quan@cs.sjtu.edu.cn}
\affiliation{%
  \institution{Shanghai Jiao Tong University}
  \city{Shanghai}
  \country{China}
}

\author{Yu Liu}
\email{ly109106@antgroup.com}
\author{Yongjun Deng}
\email{dengyongjun.dyj@antgroup.com}
\affiliation{%
  \institution{Ant Group}
  \country{China}
}

\author{Jing Yang}
\email{jyang23@gzu.edu.cn}
\author{Jiuchen Shi}
\email{shijiuchen@sjtu.edu.cn}
\author{Chen Chen}
\email{chen-chen@sjtu.edu.cn}
\author{Youmin Chen}
\email{chenyoumin@sjtu.edu.cn}
\author{Yu Feng}
\email{y-feng@sjtu.edu.cn}
\author{Minyi Guo}
\email{guo-my@cs.sjtu.edu.cn}
\affiliation{%
  \institution{Shanghai Jiao Tong University}
  \city{Shanghai}
  \country{China}
}

\begin{abstract}
As Large Language Models (LLMs) become essential in privacy-sensitive sectors like hospitals and government agencies, the on-premise LLM servers offer a cost-effective and secure alternative to public cloud services. However, these resource-constrained servers struggle to guarantee heterogeneous Service Level Objectives (SLOs) when serving multiple LoRA-adapted services simultaneously. Existing serving frameworks suffer from severe SLO violations due to the computational overhead of LoRA layers and the rigid nature of batch scheduling. To address this, we propose \sysname{}, a scheduling method tailored for LoRA-assisted on-premise LLM deployment. \sysname{} introduces two key innovations: a spatial multiplexing strategy that overlaps Base and LoRA computations by partitioning GPU Streaming Multiprocessors (SMs), and an SLO-aware scheduler that decouples request execution based on "request-level slack." By prioritizing urgent tasks and utilizing idle budget for traffic shaping, \sysname{} significantly mitigates resource contention. Our evaluation demonstrates that \sysname{} minimizes SLO violations while improving throughput compared to state-of-the-art baselines.
\end{abstract}



\maketitle

\settopmatter{printfolios=true}

\input{contents/introduction}
\input{contents/motivation.tex}
\input{contents/overview.tex}
\input{contents/searcher}
\input{contents/scheduler}
\input{contents/evaluation.tex}
\input{contents/related.tex}
\input{contents/conclusion.tex}

\begin{acks}
This work is partially sponsored by the National Natural Science Foundation of China (62302302, 62232011), and Natural Science Foundation of Shanghai Municipality (24ZR1430500). This work was supported by Ant Group through CCF-Ant Research Fund. Han Zhao is the corresponding author.
\end{acks}

\bibliographystyle{ACM-Reference-Format}
\bibliography{reference}

\end{document}

%% file: contents/environment.tex
\usepackage{booktabs}
\usepackage{outlines}
\usepackage{tikz}
\usepackage{comment}
\usepackage{amsmath}
\usepackage{caption}
\usepackage{subcaption}
\usepackage{graphicx}
\usepackage{float}
\usepackage{algorithm}
\usepackage{algorithmicx}
\usepackage[noend]{algpseudocode}
\algnewcommand{\LineComment}[1]{\State \(//\) #1}
\usepackage[normalem]{ulem}
\usepackage{bbding}
\usepackage{todonotes}
\usepackage{enumitem}
\usepackage[bottom]{footmisc}
\usepackage{hyperref}
\usepackage{cleveref}
\usepackage{pifont}
\usepackage{multirow}
\usepackage{xcolor}
\usepackage{lipsum}
\usepackage{array}
\newcolumntype{P}[1]{>{\centering\arraybackslash}p{#1}}

\definecolor{LightGray}{gray}{0.9}
\definecolor{FadedBanana}{RGB}{255,255,191}
\definecolor{DeepChalk}{RGB}{255,191,191}
\definecolor{FadedFlora}{RGB}{191,255,191}
\definecolor{DeepSnow}{RGB}{191,255,255}
\definecolor{SoapStone}{RGB}{218,218,218}
\definecolor{LightCayenneSixty}{RGB}{239,206,211}
\definecolor{LightCayenne}{RGB}{208,143,145}

\newcommand{\sysname}{{HALO}}

%% file: contents/introduction.tex
\section{Introduction}

As LLMs achieve state-of-the-art performance in many domains\cite{openai2023gpt4, deepseekai2025deepseekr1incentivizingreasoningcapability, touvron2023llamaopenefficientfoundation}, the on-premise LLM deployment has become the preferred choice for sensitive sectors like hospital and government\cite{aireport, dell, friha2024llm, unell2025real}. Compared to public LLM services like ChatGPT, the on-premise deployment offers two advantages: cost efficiency and data privacy. For instance, a single server equipped with four NVIDIA A100 GPUs costs about \$150,000. In contrast, consuming the same volume of tokens via a public service can cost about \$1,500,000 per year\cite{llmcost, pan2025cost}. More importantly, using LLM services have risks of exposing proprietary data to third-party providers.

\begin{figure}
\centering
\includegraphics[width=0.8\linewidth]{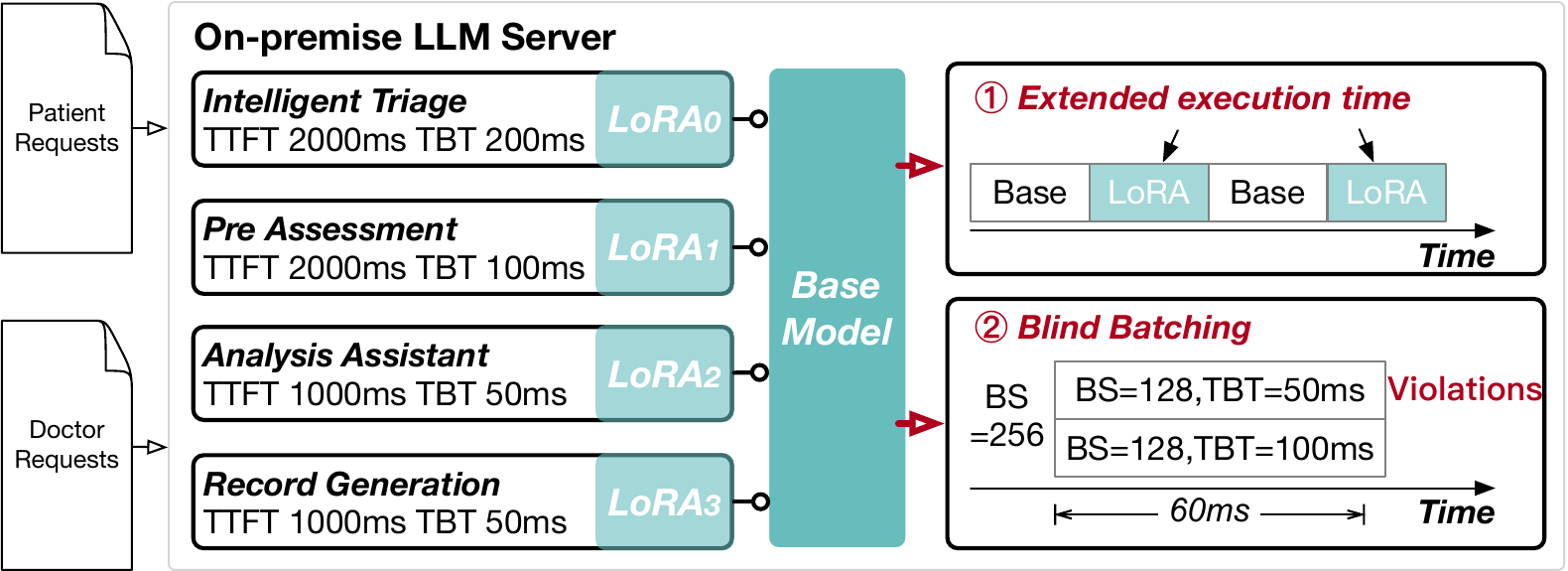}
\caption{The hospital workflow of an on-premise LLM server.}
\label{fig:appliance}
\end{figure}

\autoref{fig:appliance} shows a hospital workflow served by such an on-premise LLM server. The server hosts four services, which share a frozen base model and use Low-Rank Adaptation (LoRA)\cite{hu2022lora} for task-specific optimization\cite{slora, edgelora, punica, mobilora}. These LoRA services have different Service Level Objectives (SLOs). For example, the Analysis Assistant requires a strict Time-to-First-Token (TTFT) of 1000ms for the prefill phase and Time-Between-Tokens (TBT) of 50ms for the decode phase. In contrast, the Intelligent Triage tolerates a looser latency budget (2000ms TTFT, 200ms TBT).

Given the strict SLOs inherent to these workloads, efficient scheduling is critical. However, current on-premise LLM servers rely on the default scheduling policies of mainstream frameworks (vLLM\cite{kwon2023efficient} and SGLang\cite{zheng2023efficiently}), which lack support for multi-SLO scheduling. Consequently, we find that on-premise LLM servers are highly susceptible to request overload within limited resources. This manifests as severe SLO violations and low system throughput.
These performance issues stem from two fundamental causes: LoRA-induced throughput degradation and SLO violations driven by blind batching.

\textbf{LoRA-induced throughput degradation.}
While existing frameworks adopt the chunked-prefill mode to improve system throughput, serving LoRAs still introduces non-negligible overhead. Specifically, chunked-prefill splits prefill into smaller chunks. Each chunk merges with the decode batch to form a mixed batch, improving the overall throughput. However, LoRA adds adapters on top of the model backbone and increases computation for both prefill and decode. Our results in \S\ref{sec:causes} show that LoRAs bring an average 44.6\% throughput drop. On resource-constrained on-premise hardware, this overhead intensifies resource contention and elevates SLO violations.

\textbf{SLO-unaware blind batching.} While batching is essential for GPU efficiency, it forces all requests in a batch to advance in lockstep regardless of their individual SLOs. For example, serving Llama3-8B\cite{llama3} with LoRAs on an A100 GPU yields a decode latency of 47ms at batch size 384 and 57ms at batch size 512. If we set the maximum decode batch size for 50ms targets, 100ms-target requests in the decode batch easily block newly arrived 50ms-target requests from entering computation due to the limited request slots. This further causes TTFT violations. If we set the maximum batch size for 100ms requests, the 50ms target requests directly suffer TBT violations.

Although many works attempt to optimize SLO guarantees for LLM serving, they are insufficient for the constraints of on-premise servers. Specifically, Punica\cite{punica} reduces LoRA execution time via custom kernels, but it still suffers from sequential execution of LoRA adapters and base model operators. Meanwhile, VTC\cite{vtc} and QoServe\cite{qoserve} focus on adjusting request serving order for inter-service fairness and SLO guarantees. Once a request enters computation, it does not leave the running batch. Therefore, no existing method addresses both the throughput degradation caused by LoRA and the SLO violations caused by blind batching.

Under these circumstances, we identify two opportunities to improve the SLO guarantee in the on-premise LLM server. First, we could mitigate LoRA overhead via spatial multiplexing. By partitioning Streaming Multiprocessors (SMs) between frozen backbone (base layer) and LoRA layer, we could enable the concurrent execution of two layers. This further helps improve hardware utilization and end-to-end latency.

Second, heterogeneous SLOs provide a resource dimension overlooked by existing work: temporal slack. Requests with relaxed SLOs such as TBT of 100ms have headroom relative to their deadlines in each decode iteration. When the actual decode latency is 50ms, a request with a 100ms TBT target can execute every two iterations without violating its SLO. This means that by breaking blind batching, requests with slack can yield compute resources to requests with urgent targets or those facing SLO violations. This helps on-premise LLM serving increase acceptable concurrency and heterogeneous SLO guarantees under resource constraints.

However, capitalizing on these opportunities is non-trivial. First, determining the optimal SM partition involves a complex configuration space. The ideal split is highly sensitive to batch size, LoRA injection points (e.g., QKV vs. FFN), and hardware characteristics. This makes finding the ideal partition difficult. Second, as a request accumulates more slack, it could better improve concurrency. But once it exits, its accumulated slack drops to zero immediately. How to coordinate slack accumulation with attained throughput is a challenging problem. Furthermore, LLM inference has two stages. How to coordinate slack management across both stages to maximize system performance is another challenge.

To this end, we propose \sysname{}, a scheduling method for LoRA-assisted on-premise LLM servers. \sysname{} consists of a lightweight partition searcher and an SLO-aware stage-unified scheduler. The lightweight partition searcher finds the optimal SM partition for each base-LoRA layer pair with minimal overhead. The SLO-aware stage-unified scheduler jointly manages slack across both stages. It flexibly adjusts the compute allocation between decode and prefill within each decode iteration to maximize SLO guarantees and system throughput.

Specifically, during the serving framework initialization, the searcher automatically identifies the optimal SM partition for each base-LoRA layer pair under current hardware. To navigate the large configuration space, the searcher proposes a feedback-driven exploration strategy that minimizes the overhead. At runtime, the scheduler introduces a decoupling mechanism to maximize efficiency. It maintains an output token queue that releases generated tokens to the client precisely at their TTFT and TBT targets. This mechanism effectively acts as a traffic shaper, smoothing out bursts by buffering tokens generated during low-load periods and releasing them steadily at a high load.

Further, the scheduler introduces a novel metric called request-level slack. It unifies slack from prefill and decode and consolidates fragmented slack into longer time slices. By monitoring this slack, the scheduler identifies requests that can bypass some decode iterations without violating TBT targets. This helps prioritize urgent tasks facing SLO violations. Specifically, these urgent requests may be in the prefill stage or the decode stage. The scheduler jointly manages request scheduling across both stages to support both TTFT and TBT guarantees.

Our evaluation demonstrates that \sysname{} effectively resolves resource contention in on-premise servers. Compared to state-of-the-art baselines, \sysname{} improves P99 TTFT (99\%-ile TTFT) by 2.1$\times$ to 5.0$\times$, and achieves 19.5\% to 29.5\% throughput improvement. The main contributions of this paper are as follows:

\begin{itemize}[itemsep=2pt, parsep=2pt, labelsep=5pt, leftmargin=*, topsep=2pt,partopsep=2pt]
    \item \textbf{Characterization of Resource Contention.} We identify that severe SLO violations in on-premise LLM servers stem from two primary sources: the computational overhead of LoRA layers and the blind batching mechanism.
    \item \textbf{Automated Spatial Multiplexing.} We propose a lightweight partition searcher that spatially partitions SMs between Base and LoRA layers. This enables concurrent execution that effectively hides LoRA latency overhead without manual tuning.
    \item \textbf{SLO-aware Stage-unified Scheduling.} We design a scheduler based on a novel request-level slack metric. This decouples request execution from batch processing, allowing the system to dynamically prioritize urgent tasks and utilize the latency budget of non-urgent requests.
\end{itemize}

%% file: contents/motivation.tex
\section{Background and Motivation\label{sec:moti}}

In this section, we first provide background on LLM serving and the specific operational workflows of on-premise LLM servers. We then characterize the inefficiencies of state-of-the-art methods and identify key opportunities to address these challenges.

\subsection{LLM Basics}

\subsubsection{LLM Model Structure}

\begin{figure}
\centering
\includegraphics[width=0.8\linewidth]{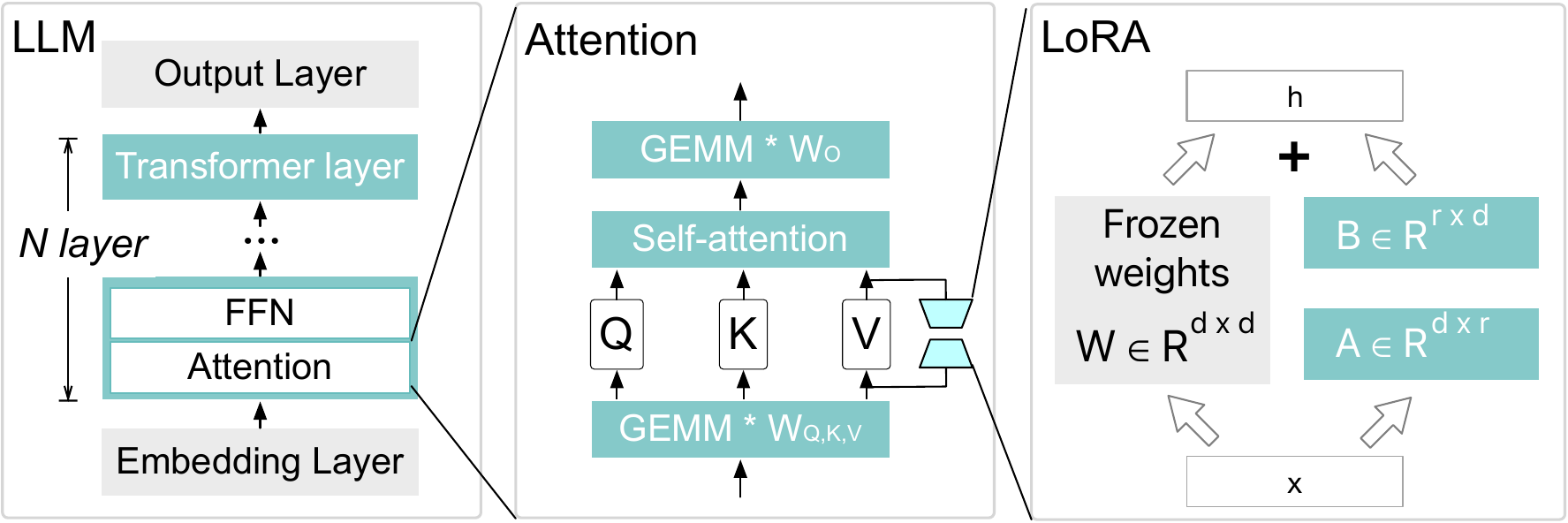}
\caption{The general architecture of an LLM model.}
\vspace{-2mm}
\label{fig:llm}
\end{figure}

Modern large language models (LLMs), such as Llama\cite{llama3} and Qwen\cite{team2024qwen2}, are primarily built upon the decode-only Transformer architecture. As shown in \autoref{fig:llm}, these models consist of a stack of identical transformer layers. Each transformer layer comprises two primary sub-layers: Self-Attention and a Feed-Forward Network (FFN).

In the attention layer, input hidden states are projected into Query, Key, and Value vectors via weight matrices ($W_Q$, $W_K$, $W_V$). The mechanism then computes attention scores to capture contextual dependencies between tokens, aggregating the results via an output projection ($W_O$). Subsequently, the FFN applies non-linear transformations, typically implemented as expansion (Gate/Up) and reduction (Down) projections.

\subsubsection{LLM Inference}

LLM inference consists of two phases: prefill and decode. The prefill phase processes the input prompt to generate the first token, while the decode phase generates subsequent tokens autoregressively. Performance is evaluated against two SLOs: Time-To-First-Token (TTFT) for the prefill and Time-Between-Tokens (TBT) for the decode.

While prefill is compute-bound, decode is memory-bound, easily suffering from low GPU utilization. To mitigate this, current systems (vLLM\cite{kwon2023efficient}/SGLang\cite{zheng2023efficiently}) adopt chunked-prefill~\cite{chunked} as the default serving mode. It splits the prefill requests into smaller chunks. Each chunk merges with the decode batch to form a mixed batch for improving the GPU utilization and overall throughput.

\subsubsection{Low-Rank Adaptation (LoRA)}

Low-Rank Adaptation is a parameter-efficient fine-tuning technique designed to adapt pre-trained LLMs to specific downstream tasks \cite{hu2022lora}. As shown in \autoref{fig:llm}, instead of updating the massive weight matrix $W \in \mathbb{R}^{d \times d}$, LoRA freezes $W$ and introduces two low-rank matrices, $A \in \mathbb{R}^{d \times r}$ and $B \in \mathbb{R}^{r \times d}$. The rank $r$ is chosen to be significantly smaller than the weight dimensions $d$. The forward pass is modified such that the output is $h = Wx + BAx$. When serving a single LoRA, we can merge $BA$ into $W$ beforehand. However, when serving multiple LoRA adapters, weight merging is infeasible. The term $BAx$ must be computed explicitly.

\subsection{Why On-premise Deployment?}

Benefitting from the rapidly improving capabilities of open-source models, on-premise LLM deployment has attracted significant industry attention. According to the IDC report\cite{idc}, on-premise infrastructure accounted for 15.9\% of global AI infrastructure spending in Q2 2025. This trend is rooted in two advantages: absolute data privacy and low operating costs.

\textbf{Data privacy.} On-premise deployment ensures that sensitive data never leaves the organization's own infrastructure, which is critical for hospitals, government agencies, and enterprises handling proprietary information.

\textbf{Cost efficiency.} Consider a mid-sized hospital running an on-premise LLM server at 15 requests/second for 12 hours/day, yielding roughly 648,000 requests/day. Assuming an average input length of 512 tokens and output length of 256 tokens, the annual token volume reaches 121B input tokens and 60.6B output tokens. At GPT-4o pricing (\$2.50/1M input tokens, \$10.00/1M output tokens), the equivalent public cloud cost totals \$900,000 per year. By contrast, a single 4×A100 server can sustain this workload (RPS$\approx$15) at a one-time hardware cost of only \$150,000, which is less than one quarter of the annual cloud expenditure.

\subsection{Multi-SLOs in On-premise LLM Servers}

\begin{table}
\centering
\small
\caption{The representative LLM services in a hospital and their corresponding SLOs and datasets. TBT-1 is the TBT target for 7B and 8B LLMs, while TBT-2 is the TBT target for 30B LLMs. Following prior works \cite{chunked, mooncake}, 30B LLMs have looser TBT targets.}
\vspace{-2mm}
\begin{tabular}{l||l|l|l|l}
\toprule
\textbf{Services}          & \textbf{TTFT} & \textbf{TBT-1} & \textbf{TBT-2} & \textbf{Dataset}             \\\hline
Intelligent Triage& 2000ms& 200ms& 200ms& ShareGPT \cite{sharegpt} \\\hline
Pre Assessment    & 2000ms& 100ms& 200ms& ChatDoctor \cite{chatdoctor} \\\hline
Analysis Assistant& 1000ms& 50ms & 100ms & AlpaCare \cite{alpacare} \\\hline
Record Generation & 1000ms& 50ms &  100ms & MedicalData \cite{medicaldata} \\
\bottomrule
\end{tabular}
\label{tab:hospital}
\end{table}

\begin{table}
\centering
\small
\caption{The representative LLM services in a service center and their corresponding SLOs and datasets.}
\vspace{-2mm}
\begin{tabular}{l||l|l|l|l}
\toprule
\textbf{Services} & \textbf{TTFT} & \textbf{TBT-1} & \textbf{TBT-2} & \textbf{Dataset}             \\\hline
Service Triage& 2000ms & 200ms & 200ms & ShareGPT  \cite{sharegpt} \\\hline
Document Check    & 3000ms & 100ms & 200ms & Alpaca \cite{alpaca} \\\hline
Realtime Translation& 1000ms & 50ms & 100ms & WMT \cite{wmt} \\\hline
Information Retrieval & 1000ms & 50ms & 100ms & AzureTrace \cite{splitwise} \\
\bottomrule
\end{tabular}
\label{tab:center}
\end{table}

The on-premise LLM servers have become the preferred deployment solution for privacy-sensitive sectors \cite{aireport, dell, friha2024llm, unell2025real}. \autoref{tab:hospital} and \autoref{tab:center} illustrate representative services found in a hospital and a government service center.

As for the hospital scenario, there are four mainstream services: {\it Intelligent Triage}, {\it Pre Assessment}, {\it Analysis Assistant}, and {\it Record Generation}. The workflow follows the patient journey: a patient first interacts with the intelligent triage agent for routing, followed by a detailed pre-consultation assessment. Subsequently, the doctor leverages the analysis assistant for real-time diagnostic support and concludes by using the medical record generation.

As for the service center scenario, there are also four mainstream services: {\it Service Triage}, {\it Document Check}, {\it Realtime Translation}, and {\it Information Retrieval}. Citizens are first routed by the service triage agent, followed by document pre-screening to verify materials. Finally, clerks utilize translation and retrieval tools to facilitate the transaction.

Crucially, as human experts (doctors and clerks) constitute the critical path in these interactions, the services assisting them demand significantly stricter latency guarantees. As shown in \autoref{tab:hospital}, the {\it Analysis Assistant} requires a strict TBT of 50ms (with a 1000ms TTFT) to ensure conversational fluidity, whereas the automated {\it Intelligent Triage} tolerates a looser 200ms TBT (2000ms TTFT) \cite{vtc, sloserve, adaserve}. To evaluate these scenarios realistically, we utilize datasets derived from actual medical and administrative scenarios.

\subsection{Inefficiencies}

Since existing frameworks lack support for multi-SLO guarantees, current on-premise deployments set the maximum running batch size based on the strictest TBT target. Under the chunked-prefill mode, this maximum batch size limits the total size of the mixed batch (decode batch size + prefill chunk size). Meanwhile, because chunked-prefill prioritizes TBT over TTFT, existing systems rarely encounter TBT violations but mainly suffer from TTFT violations.

\subsubsection{SLO violations}

In this experiment, we configure the LLM server with a single NVIDIA A100 GPU to host the four hospital services. The system serves Llama3-8B using SGLang\cite{zheng2023efficiently} in chunked-prefill mode. The maximum batch size is set to 384 based on the 50ms TBT target.

\begin{figure}
\centering
\includegraphics[width=\linewidth]{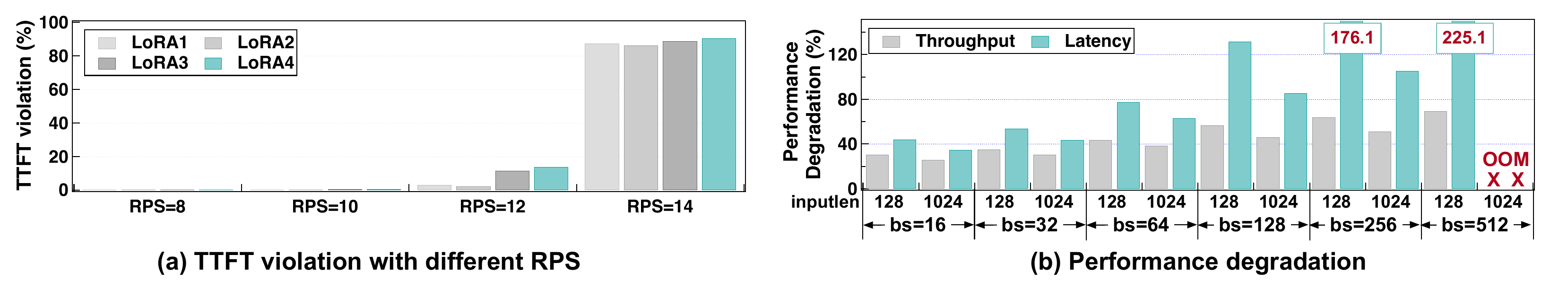}
\vspace{-4mm}
\caption{(a) The TTFT violation of the current serving system under different workloads (requests per second, RPS). L1, L2, L3, and L4 represent four LoRA services, which are  {\it Intelligent Triage}, {\it Pre Assessment}, {\it Analysis Assistant}, and {\it Record Generation}. (b) The performance degradation of decode latency and throughput under different running setups.}
\label{fig:moti-slo-1}
\end{figure}

\autoref{fig:moti-slo-1}(a) shows the TTFT violation rates across varying load levels (requests per second, RPS). As shown, services with strict SLOs begin to violate deadlines when the load exceeds 10 requests per second. After the load passes 12 requests per second, all services suffer significant violations. This demonstrates that the scheduling policies of existing frameworks fail to maintain SLOs under limited compute resources.

\subsubsection{Root Causes}
\label{sec:causes}

Our in-depth analysis attributes these SLO violations to two primary inefficiencies: sequential and inefficient LoRA execution, and SLO-agnostic blind batching.

Firstly, we collect the latency and throughput performance with four LoRA services enabled. Our results show that LoRA-enabled inference suffers an average 94.5\% latency increase and 44.6\% throughput degradation (Input Length: 128 and 1024, Output Length: 8). This is because the base layer and LoRA layer execute sequentially even though they have no data dependency. To address this problem, Punica employs specialized kernels to accelerate LoRA execution. \autoref{fig:moti-slo-1}(b) shows the normalized performance of LoRA-enabled serving compared to the vanilla baseline. It still incurs an average 31.7\% latency degradation and 23.8\% throughput degradation. This indicates that LoRA serving with Punica\cite{punica} still suffers from significant performance loss.

\begin{figure}
\centering
\includegraphics[width=\linewidth]{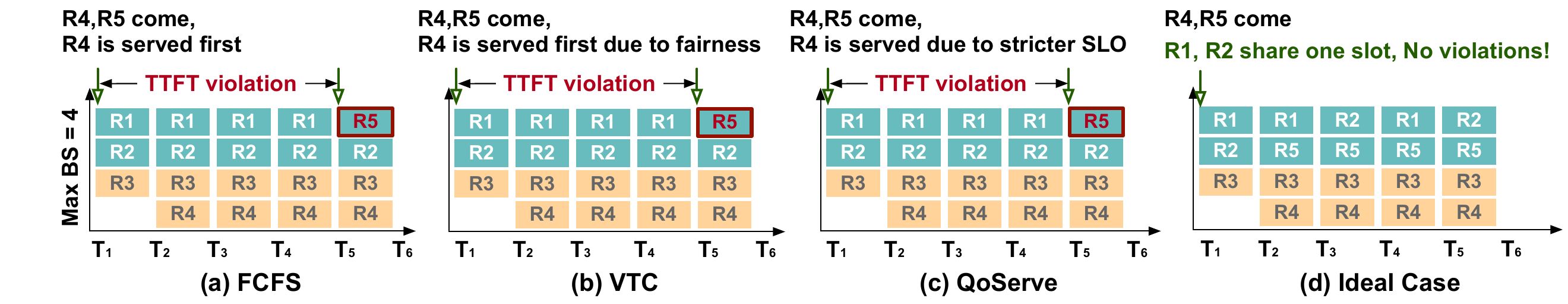}
\vspace{-4mm}
\caption{The scheduling decisions and effects under FCFS, VTC, QoServe and ideal case.}
\label{fig:moti}
\vspace{-2mm}
\end{figure}

Secondly, we show why existing systems suffer from SLO violations and why existing methods cannot solve this problem. \autoref{fig:moti} shows the scheduling effects of the FCFS method and two state-of-the-art scheduling methods, VTC\cite{vtc} and QoServe\cite{qoserve}. Specifically, VTC adjusts the request serving order to guarantee inter service fairness. This means multiple services process a fair number of tokens within a given time period. QoServe attempts to prioritize user-facing services and defer low-priority offline inference to optimize SLO guarantees. In addition, QoServe dynamically adjusts the maximum batch size based on the strictest user-facing service TBT target in the running batch to improve GPU utilization.

As shown in \autoref{fig:moti}, assume the maximum batch size on the current GPU is 4. The base time unit between two consecutive timestamps is 50ms. {\it R1}, {\it R2}, and {\it R3} are currently in computation. {\it R1} and {\it R2} belong to service {\it LoRA1} with a TTFT target of 200ms and a TBT target of 100ms. {\it R3} belongs to service {\it LoRA2} with a TTFT target of 100ms and a TBT target of 50ms. At time $T_1$, requests {\it R4} and {\it R5} arrive. {\it R4} belongs to service {\it LoRA2} and {\it R5 belongs} to service {\it LoRA1}.

For the FCFS method, when {\it R4} and {\it R5} arrive at time $T_1$, only one running slot remains. {\it R4} is served while {\it R5} waits in the queue. Since LLM requests occupy a slot for an extended period due to the autoregressive nature of decoding, {\it R5} suffers a TTFT violation at time $T_4$.

For the VTC method, since the dominant service in the current system is service {\it LoRA1}, the fairness calculation determines that {\it R4} from service {\it LoRA2} should be prioritized over {\it R5}. After {\it R4} is served, {\it R5} still suffers a TTFT violation at time $T_4$.

For the QoServe method, it adopts earliest deadline first strategy to schedule user facing services under high load. Since {\it R4} has a tighter TTFT target than {\it R5}, {\it R4} is served first. As a result, {\it R5} still suffers a TTFT violation at time $T_4$. Note that, since all requests belong to user-facing services, the offline inference deferral proposed by QoServe does not apply in the on-premise scenario.

In the ideal case, when {\it R4} and {\it R5} arrive, {\it R4} is served first. Since {\it R1} and {\it R2} both have a TBT target of 100ms and the current decode iteration takes 50ms, {\it R1} and {\it R2} can share one running slot by executing alternately. This frees up a slot so that {\it R5} can be served immediately. In this way, the TTFT violation for {\it R5} is eliminated.

Based on the above analysis, existing methods fail to provide SLO guarantees due to blind batching. Once a request enters the running batch, it cannot exit. Even if it has a relaxed SLO and could yield its running slot, it cannot help requests with urgent SLOs avoid violations.

\subsection{Opportunities and Challenges}

In a nutshell, while existing serving frameworks simplify multi-service deployment through batching, they suffer from the throughput degradation inherent to LoRA computation and the rigidity of SLO-agnostic scheduling. Facing these limitations on resource-constrained on-premise LLM servers, we identify two major opportunities to improve performance.

\begin{itemize}[itemsep=2pt, parsep=2pt, labelsep=5pt, leftmargin=*, topsep=2pt,partopsep=2pt]
    \item We can reduce LoRA overhead using spatial multiplexing. By splitting the SMs between the base layer and the LoRA layer, we can run both components simultaneously. This parallelism aligns their completion times and improves the latency.
    \item Heterogeneous SLOs provide a resource dimension overlooked by existing work: temporal slack. Requests with relaxed SLOs have headroom relative to their deadlines in each decode iteration. This slack can be used to increase the concurrency capacity and improve overall SLO guarantees.
\end{itemize}

However, utilizing these two opportunities is non-trivial. First, implementing spatial multiplexing requires finding the optimal SM partition from a great search space. Finding the optimal one is challenging as it changes based on batch size and the specific LoRA layers used (e.g., QKV vs. FFN). Second, slack accumulation does not necessarily translate into throughput improvement. Once a request exits, its accumulated slack becomes meaningless. How to coordinate slack accumulation with attained throughput is a challenging problem. Furthermore, how to coordinate slack management across both stages to maximize system performance is another challenge.

%% file: contents/overview.tex
\section{Design Overview}
\label{sec:overview}

\begin{figure}
\centering
\includegraphics[width=0.8\linewidth]{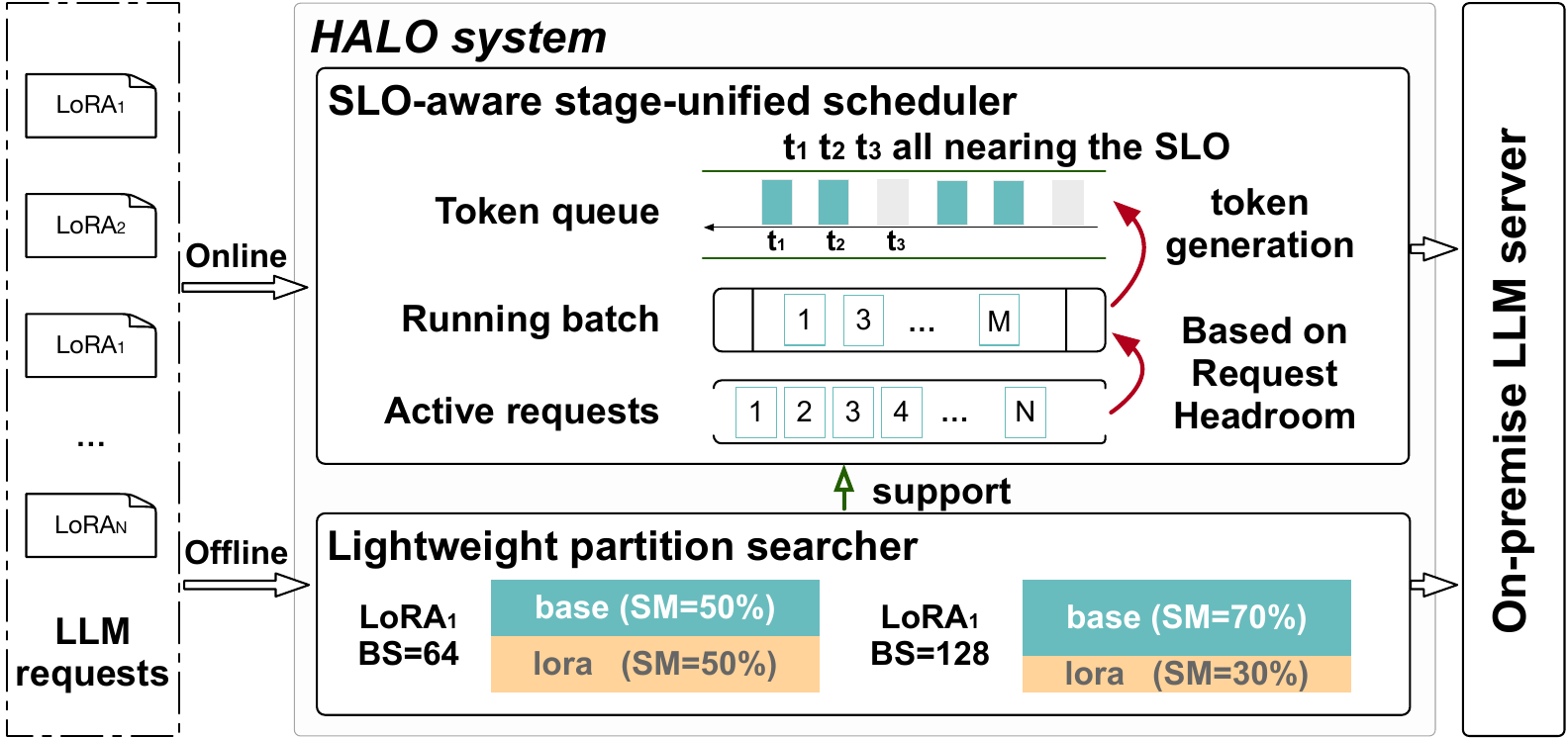}
\caption{The design overview of \sysname{}.}
\label{fig:halo}
\end{figure}

In this section, we present \sysname{}, a scheduling method designed to enforce heterogeneous SLOs on resource-constrained on-premise LLM servers. As shown in \autoref{fig:halo}, \sysname{} consists of two core components: a lightweight partition searcher and an SLO-aware stage-unified scheduler. The partition searcher addresses LoRA induced throughput degradation by finding the optimal SM partition for concurrent base-LoRA execution. The scheduler addresses SLO violations by exploiting temporal slack from heterogeneous SLOs. Specifically, it jointly manages request scheduling across both the prefill and decode stages to maximize SLO guarantees and system throughput.

The workflow begins with the offline partition searcher, which automatically locates the optimal SM partition for each base-lora layer pair at different batch sizes. Since the optimal configuration varies significantly with LoRA injection points (e.g., QKV vs. FFN), batch size, and hardware characteristics, exhaustive search is infeasible. The searcher first automatically profiles the independent execution time of two layers at different SM ratios. Then, it utilizes a feedback-driven profiling method to locate the optimal resource configuration for each base-lora layer pair.

At runtime, the SLO-aware stage-unified scheduler replaces default First-Come-First-Served policies with a deadline-aware mechanism. Upon arrival, each request is tagged with its specific SLO targets (TTFT and TBT). The scheduler continuously calculates a novel metric called request-level slack. This slack quantifies the dynamic headroom between the current time and the strict deadline of the next token to be generated. This metric unifies slack from both the prefill and decode stages, consolidating fragmented slack into actionable time slices. Requests with small slack are flagged as urgent for execution, while those with ample slack are candidates for deferral.

Leveraging this analysis, the scheduler performs decoupled batch formation. It breaks the "lockstep" constraint of traditional batching by constructing execution batches that include only urgent requests and skip non-urgent ones. This allows requests with slack to yield their compute slots to urgent requests facing imminent TTFT or TBT violations in either the prefill or decode stage. By jointly managing scheduling across both stages, the scheduler flexibly adjusts the compute allocation between decode and prefill within each iteration to maximize SLO guarantees.

Finally, the output token queue acts as a traffic shaper to decouple variable decode latency from the user experience. Generated tokens are buffered and released to the client strictly according to their TBT intervals. This smooths execution jitter, ensuring a steady stream of text. Crucially, this mechanism allows the system to "bank" time credits; if tokens are generated faster than required, the queue holds them, building a buffer that enables the scheduler to absorb future high-priority traffic bursts without violating SLOs.

Note that the base-LoRA parallelism improves performance only during the decode phase and has no effect for the prefill, since LoRA computation during prefill already achieves high utilization.

%% file: contents/searcher.tex
\section{Lightweight Partition Searcher}

\subsection{Requirements and Challenges}

\subsubsection{Requirements}

While leveraging base-LoRA layer parallelism could improve system throughput, it imposes two critical requirements on the serving system. First, the system must be capable of identifying the optimal SM partition configuration for arbitrary models under varying execution contexts. Second, the configuration search process needs to run automatically during framework initialization, which makes it essential to use an exploration method with very low overhead so that startup latency remains minimal.

\subsubsection{Challenges}

However, satisfying the above two requirements presents a non-trivial challenge due to the combinatorial explosion of the configuration space.

First, current LoRA services insert adapters to all possible positions (QKV, O, Gate/Up, Down), but their ranks can be chosen independently\cite{edgelora, elora}. Second, modern frameworks rely on CUDA graph to accelerate execution, which requires capturing a separate CUDA graph for each batch size. As a result, the optimal SM partition must be prepared for every batch size before CUDA graph capture. Third, even for a single base-LoRA layer pair, the partition search space is already very large. For example, on an NVIDIA A100 (108 SMs) with a partition granularity of 2 SMs (via GreenContext), there are $\sum_{k=0}^{54}(55-k)=1540$ valid configurations.

Based on this analysis, we can quantify the total profiling cost in \autoref{eq:overhead1}:

\begin{equation}
\label{eq:overhead1}
    T_{total} = N_{lora} \times N_{bs} \times N_{conf} \times T_{decode}
\end{equation}

Here, $N_{lora}$ denotes the number of LoRA positions, $N_{bs}$ the number of captured batch sizes, and $N_{conf}$ the number of candidate partitions. $T_{decode}$ represents the profiling duration required to obtain stable measurements ($\approx$100ms). For a typical setup ($N_{lora}=4, N_{bs}=8, N_{conf}=1540$), the profiling time exceeds 1.3 hour per model. This prohibitive overhead renders exhaustive search infeasible for production environments.

\subsection{Feedback-based Search Method}

\subsubsection{Observations}

To reduce the overhead described above, we attempt to design a lightweight search method that can lower the overall cost. As a first step, we conduct a series of experiments to better understand the compute behavior. These experiments lead to several key observations that guide the design.

\begin{figure}
\centering
\includegraphics[width=0.8\linewidth]{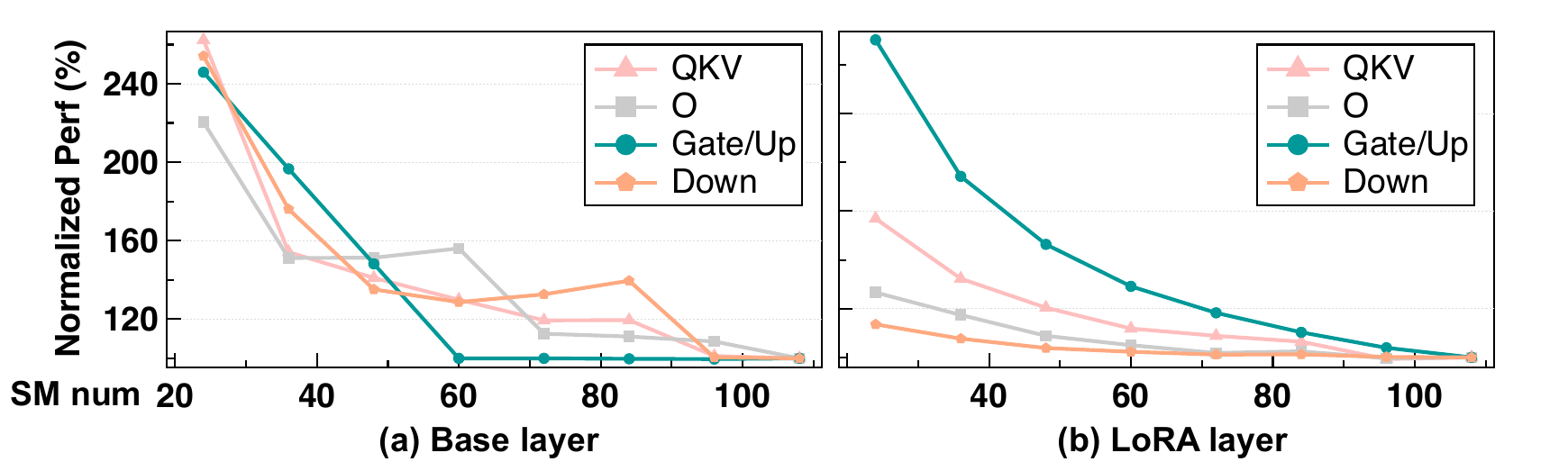}
\caption{The normalized performance of the base layer and the LoRA layer under different SM allocations.}
\label{fig:lora-perf-sm}
\end{figure}

\textbf{1. SM allocation:} \autoref{fig:lora-perf-sm} shows the performance of the base layer and the LoRA layer under different SM allocations. The data points are collected using $ncu$ for four LoRA insertion points (QKV, O, Gate/Up, Down) in Llama3-8B with a batch size of 64 on an NVIDIA A100. All results are normalized to the best performance achieved when using all SMs. As shown, the performance gains of both two layers become very small once a certain number of SMs is allocated.

This behavior occurs because the workload of an operator is determined by the number of thread blocks, which cannot always be distributed evenly across SMs. For example, if there are 200 thread blocks and 80 SMs, some SMs are assigned two thread blocks while others are assigned three, and those with three blocks determine the overall execution time. In this case, allocating 100 SMs could improve performance, whereas allocating 90 SMs would not.

\textbf{Observation-1:} We can allocate fewer SMs to the base layer and the LoRA layer while incurring only negligible performance degradation (within 5\%).

\textbf{2. SM partition:} Since the command line profiling tool $ncu$ cannot measure the execution time of two layers running in parallel, we can only rely on the overall decode latency to determine whether performance improves. However, $nsys$ is primarily designed for visual inspection and makes it difficult to obtain fine-grained, per-layer quantitative performance metrics in a form that can be automatically processed via the command line.

\begin{figure}
\centering
\includegraphics[width=0.7\linewidth]{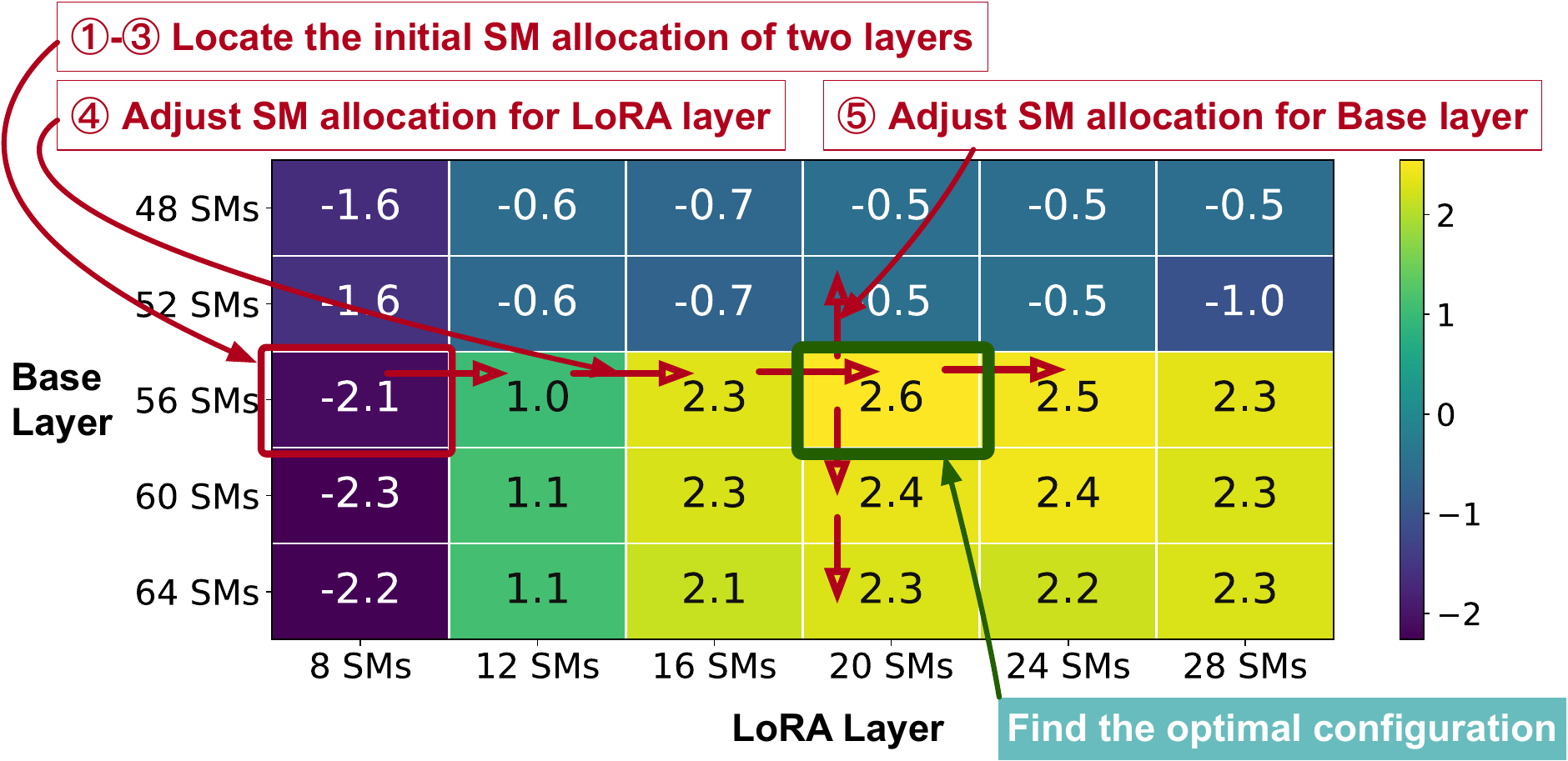}
\caption{The absolute performance gain (milliseconds) with different SM allocations.}
\vspace{-4mm}
\label{fig:parallelism}
\end{figure}

\autoref{fig:parallelism} shows the absolute latency gain (milliseconds) of Llama3-8B on an NVIDIA A100 when adjusting the SM partition between the Gate/Up layer and its LoRA layer. In this experiment, we use an input length of 128 and a batch size of 64. All data points are measured as the original sequential decode latency minus the decode latency with base-LoRA parallelism. As shown, when the Gate/Up layer is assigned 52 SMs, the decode latency even become worse. This is because the performance benefit from parallel execution is not large enough to offset the performance loss caused by reducing the resources allocated to the base layer.

\textbf{Observation-2:} The allocation of SMs in the base layer needs to be carefully considered, as its performance degradation may outweigh the advantage brought by base-LoRA parallelism.

\textbf{3. Parallel effect:} We further analyze the parallelism between the base layer and the LoRA layer in \autoref{fig:parallelism} using the $nsys$ tool. We find that none of the SM partitions in \autoref{fig:parallelism} achieve perfect overlap between the base and LoRA layers, meaning that they do not finish at the same time. When we fix the SM allocation for the base layer and vary the allocation for the LoRA layer, or fix the allocation for the LoRA layer and vary the allocation for the base layer, only two outcomes appear. Either the decoding latency improves or it worsens. When the decoding latency improves, the degree of overlap between the base and LoRA layers also improves. When the decoding latency worsens, the overlap becomes worse.

For example, when 56 SMs are assigned to the base layer, increasing the number assigned to the LoRA layer initially leads to a steady reduction in decoding latency because the overlap between the base and LoRA layers continues to improve. Once the number of SMs assigned to the LoRA layer exceeds 24, the latency starts to increase because the overlap between the base and LoRA layers begins to degrade, and this degradation becomes more pronounced as more SMs are added.

\textbf{Observation-3:} Once the SM allocation of one layer is fixed, a feedback driven search can be used to find the optimal allocation for the remaining layer.

\textbf{Observation-4:} The search problem with no prior knowledge over two variables (the SM allocation of two layers) can be decomposed into a two-step feedback-driven process.

\subsubsection{Search Algorithms}

Based on these observations, we design a feedback based search method. The method proceeds in five stages.

\begin{enumerate}[itemsep=2pt, parsep=2pt, labelsep=5pt, leftmargin=*, topsep=2pt,partopsep=2pt]
\item We use $ncu$ to statistically measure the performance curves of the base layer and the LoRA layer with different SM allocations for every batch size setting.
\item Using these performance curves, we select the SM allocation for the base layer such that its performance drops by less than 10\% from the optimum. We identify this point by collecting all performance spike points across the curves. At a spike point, such as 64 SMs for the Gate/Up layer, allocating 60 SMs leads to a performance drop of more than 10\%, while increasing SMs yields no improvement of 10\% or more. We find that these spikes have average improvement of 40.9\%, which makes this cutoff a reliable way to preserve near optimal performance.
\item We choose the smallest SM number for the LoRA layer whose computation time is close to that of the base layer. When memory contention is ignored, this choice allows the two layers to finish at roughly the same time.
\item We fix the SM allocation for the base layer and gradually increase the allocation for the LoRA layer. As long as the decoding latency continues to improve, we keep increasing it. Once a performance drop is observed, we reverse direction and reduce the allocation until we reach the best performance point.
\item After fixing the SM allocation for the LoRA layer, we apply the same procedure to the base layer. We increase its allocation while decoding latency improves, and once performance degrades, we decrease it until the optimal point is found.

\end{enumerate}

\autoref{fig:parallelism} illustrates an example of this search process. As shown in the figure, the method first identifies an initial allocation of 56 SMs and 8 SMs. It then adjusts the allocation for the base layer to refine this result, followed by adjustments to the allocation for the LoRA layer to further improve performance.

Overall, this method finds a performance point within 1\% of the optimal in 85.4\% of cases, and in the remaining cases the performance loss is within 3\%.

%% file: contents/scheduler.tex
\section{Scheduling Optimization}

In this section, we present an SLO-aware stage-unified scheduler designed to address the violations caused by blind batching and maximize SLO-compliant throughput under constrained resources.

\subsection{Current Workflow}

\begin{figure}
\centering
\includegraphics[width=\linewidth]{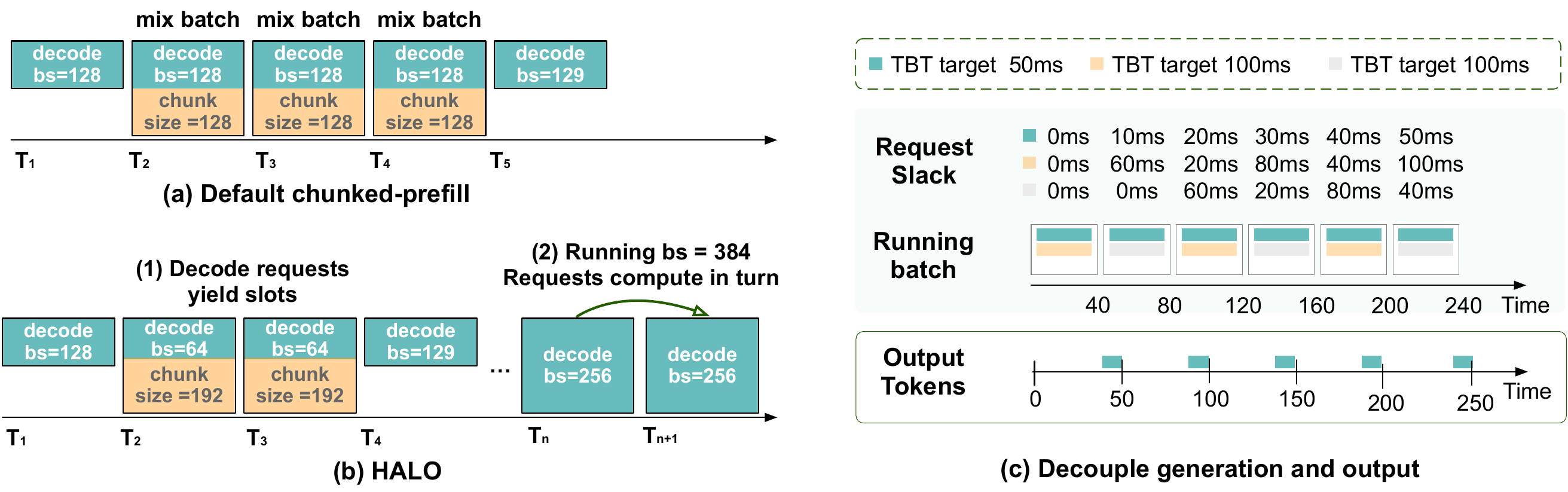}
\caption{The scheduling effect of chunked-prefill and HALO.}
\label{fig:workflow}
\end{figure}

Since the decode phase easily suffers from low GPU utilization due to its memory-bound nature, existing systems (SGLang and vLLM) adopt chunked-prefill proposed by Sarathi-serve to improve overall GPU utilization and system throughput. Specifically, when a new request arrives, its prefill computation is divided into multiple prefill chunks. These chunks merge with the decode batch to form a mixed batch for execution. Under the chunked-prefill mode, every iteration includes the decode batch. This causes the system to prioritize TBT targets over TTFT targets. As a result, existing systems generally set the maximum running batch size based on the strictest TBT target. This value limits both the mixed batch and the normal batch.

\autoref{fig:workflow}(a) shows a specific scheduling example with a maximum batch size of 256. Initially, the decode batch size is 128. When {\it R1} with an input length of 384 arrives, it is divided into multiple prefill chunks. The chunk size is calculated as max batch size minus decode batch size, which is 256 minus 128, giving 128. Therefore, the prefill computation of this request is divided into three prefill chunks. They participate in three decode iterations. After the three mixed batches complete, the serving system falls back to the normal decode batch with a batch size of 129.

From the above process, existing systems are unaware of heterogeneous SLOs across requests. This causes requests with relaxed targets to occupy valuable request slots, leading to SLO violations for requests with strict targets. To address this, we attempt to break the blind batching in existing systems and exploit slack from relaxed requests to improve overall SLO guarantees. However, this introduces several challenges.

\subsection{Challenges for Efficient Scheduling}

First, the slack of a request is generated per token, but there is not always a need to exploit it due to the fluctuated workload. However, once a request exits, its accumulated slack becomes meaningless. More importantly, directly exploiting fragmented slack introduces significant scheduling overhead. Therefore, how to efficiently utilize fragmented slack is the first challenge.

Second, each request has two stages: prefill and decode. When requests cannot enter the decode batch in time, they first suffer TTFT violations. This introduces the second challenge: how to jointly manage compute allocation across both stages. In addition, breaking blind batching also allows the system to exceed the preset maximum batch size. However, increasing running capacity without limits will eventually cause TBT violations. This introduces the third challenge: how to set the maximum running capacity to guarantee that TBT is not violated.

\subsection{Decoupling Token Generation and Token Output}

To solve the first challenge, we first decouple token generation from token output. Token generation is only responsible for producing tokens. Token output is responsible for returning tokens to the client side according to the TBT target. Further, we define a new metric for each request denoted as $Slack_{req}$. This metric represents how long a request can be deferred without participating in decoding. Every time a request completes one token computation, HALO updates the $Slack_{req}$ value for every request using \autoref{eq:headroom}. In this equation, $T_{decode}$ is the current decoding time, $T_{mix}$ is the time for the mix batch. Here $N$ denotes the number of elapsed mixed-batch iterations since the request arrived.

\begin{align}
\label{eq:headroom}
    Slack_{req} &+= (TBT - T_{decode})\ \ \ \ \ \ \ \ \ \ \ \ \ for\ Decode \\
    Slack_{req} &+= (TTFT - N * T_{mix})\ \ \ \ \ \ \ for\ Prefill
\end{align}

Based on these two designs, we can accumulate slack from relaxed requests during low load periods. When a bursty load arrives, these requests can yield compute resources for an extended period. However, when a request finishes its computation and leaves the current decode batch, its $Slack_{req}$ is reset to zero. This means the computational margin it has accumulated no longer contributes to improving overall system throughput. As a result, continuously increasing the headroom of requests with relaxed SLOs is not optimal.

To maximize the benefit of accumulated computational margin and to minimize wasted margin, we use the average difference in $Slack_{req}$ across all requests as an evaluation metric. Under this criterion, when any request exits, the wasted computational margin is minimized in the system.

\subsection{Stage-Unified Management}

After solving the first challenge, we propose stage unified management to address the second challenge. First, we derive an observation from managing the standard deviation of $Slack_{req}$ across requests. We should consider how to minimize the standard deviation of $Slack_{req}$ before a request enters computation. Therefore, when a request arrives, its initial $Slack_{req}$ is zero. It should complete its computation as quickly as possible. By finishing computation sooner, we can reach an equilibrium of $Slack_{req}$ across all requests faster.

Therefore, whenever a new request arrives, we should yield slots from requests in the current decode batch that have available slack. These requests stop computation temporarily to help the new request finish its prefill computation as quickly as possible. For example, when multiple requests have accumulated slack of N decode iterations, they should yield their request slots for the next N iterations to accelerate prefill completion.

Meanwhile, breaking blind batching removes the default maximum batch size constraint. However, continuously increasing running capacity will eventually cause TBT violations. To avoid TBT violations, similar to the default scheduling policy, we need to calculate the maximum capacity under the current scheduling method. We find that we can compute the actual capacity usage based on TBT targets. For example, if the strictest TBT target is 50ms, a request with a 100ms target only participates in computation once every two iterations. This means its actual capacity usage is 0.5.

Based on this calculation method, we can compute the actual capacity usage of all requests and ensure it remains below the originally set maximum batch size. This guarantees that \sysname{} does not violate TBT targets. Finally, we apply the EDF (earliest deadline first) strategy to sort queued requests. This further reduces TTFT violations.

%% file: contents/evaluation.tex
\section{Evaluation}

\subsection{Experiment Setup}

\begin{table}[]
\centering
\small
\caption{The models and corresponding GPU usage.}
\begin{tabular}{l|c|c|c|c}
\toprule
Model & Llama3-8B      & Yi1.5-34B      & Qwen2.5-7B     & Qwen2.5-32B    \\
\hline
GPU        & A100 (80G) * 1 & A100 (80G) * 4 & A100 (80G) * 1 & A100 (80G) * 4 \\
\bottomrule
\end{tabular}
\label{tb:model}
\end{table}

\begin{table}[t]
\centering
\small
\caption{The hardware and software specifications.}
\label{tb:configuration}
\resizebox{\textwidth}{!}{
\begin{tabular}{c|c|c|c|c|c}
\toprule
CPU & GPU & OS & Driver & CUDA Runtime & Serving Framework \\
\hline
AMD EPYC 7513 & A100, Ada6000 & Ubuntu 22.04 & 570.144 & CUDA 12.9 & SGLang v0.5.4 \\
\bottomrule
\end{tabular}
}
\end{table}

\subsubsection{Testbed}
We evaluate \sysname{} on large language models of different scales under both single GPU and multi GPU settings to demonstrate its effectiveness and generality. \autoref{tb:model} summarizes the evaluated models and their corresponding hardware configurations. As shown, we use a single A100 GPU to serve Llama3-8B\cite{llama3} and Qwen2.5-7B\cite{team2024qwen2}. For larger models, including Yi1.5-34B\cite{ai2024yi} and Qwen2.5-32B\cite{team2024qwen2}, we use a server equipped with four A100 SXM4 GPUs with 80 GB of memory each. These GPUs are connected through NVLink. \autoref{tb:configuration} lists the software configurations used in our experiments, which are kept consistent across all models.

Note that, Yi1.5-34B\cite{ai2024yi} is built on the Llama model architecture as there is no official Llama3-30B. In addition, we also evaluate the \sysname{} on the Nvidia Ada6000 to demonstrate its effectiveness.

\subsubsection{Workloads}
\begin{figure*}
\centering
\includegraphics[width=0.8\columnwidth]{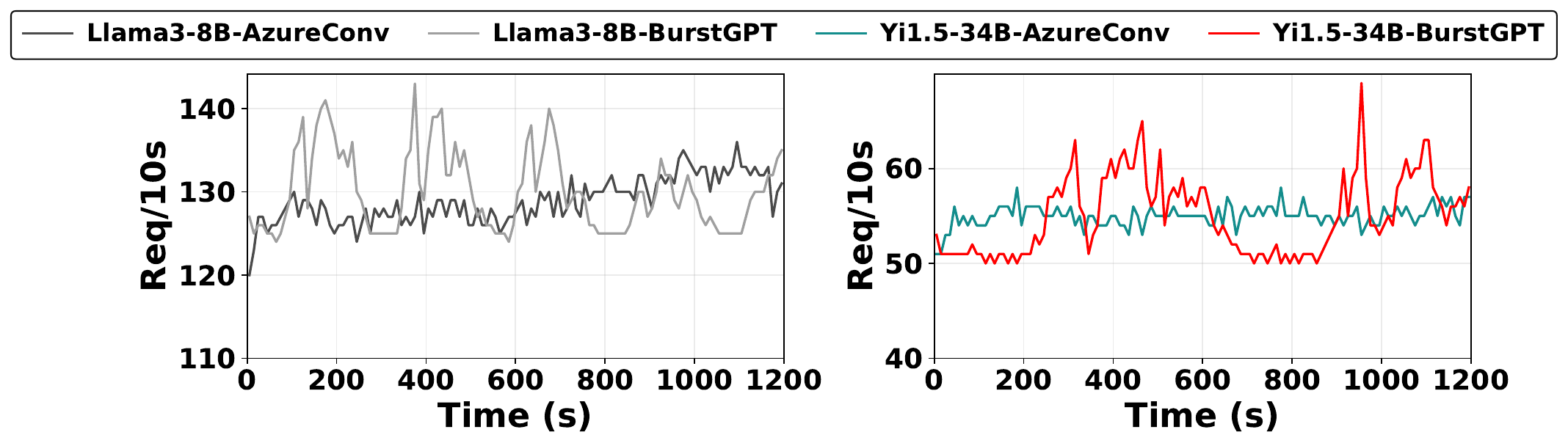}
\caption{The two real-world workload traces after scaling in hospital Scenario.
}
\label{fig:exp6-1-1}
\end{figure*}

Following prior works\cite{wu2024dlora,slora}, we generate workloads using request patterns derived from production traces, since these traces reflect real world LLM usage. At the same time, we used the input length and output length distributions from the dataset, because the original LLM traces are not collected from an on-premise deployment scenario.

\subsubsection{Metrics}

HALO focuses on system throughput under guaranteed SLOs. Following prior works\cite{chunked, splitwise, mooncake, zhong2024distserve}, we use tail latency such as P99 to assess SLO attainment. Meanwhile, there is another metric for the decode phase: TPOT (time per output token). In comparison, TBT accounts for the latency of each individual token, whereas TPOT is an average metric that may mask poor performance of some tokens [42]. Thus, we choose TBT over TPOT as a stricter SLO metric. \autoref{tab:hospital} and \autoref{tab:center} show all SLO setups for all models.

\subsubsection{Baselines}
We compare \sysname{} with four baselines. All baselines run in the chunked-prefill mode proposed by Sarathi-Serve\cite{chunked}.
\begin{itemize}
[leftmargin=*,topsep=0.2em]
    \item {\bf SLoRA:} SLoRA\cite{slora} is a  state-of-the-art multi-LoRA serving system that batches the computation of multiple LoRA adapters and uses a first come first served (FCFS) scheduling policy.
    \item {\bf VTC-T:} VTC\cite{vtc} targets fairness in multi-tenant LLM serving based on token-level metrics. We use VTC-T to denote the original VTC design. It enforces fairness across four services by equalizing the number of generated tokens.
    \item {\bf VTC-R:} VTC-R is a VTC variant that enforces fairness using the per-service TTFT (first-token) violation rate, defined as the number of requests whose first token misses the TTFT target divided by the total number of requests of that service.
    \item {\bf QoServe:} QoServe\cite{qoserve} is another state-of-the-art multi-LoRA serving system. It adopts the earliest deadline first strategy under high load and dynamically adjusts the chunk size based on the strictest TBT target in the batch.
\end{itemize}

\subsection{Evaluation in Hospital Scenario}

\begin{figure*}
\centering
\includegraphics[width=\columnwidth]{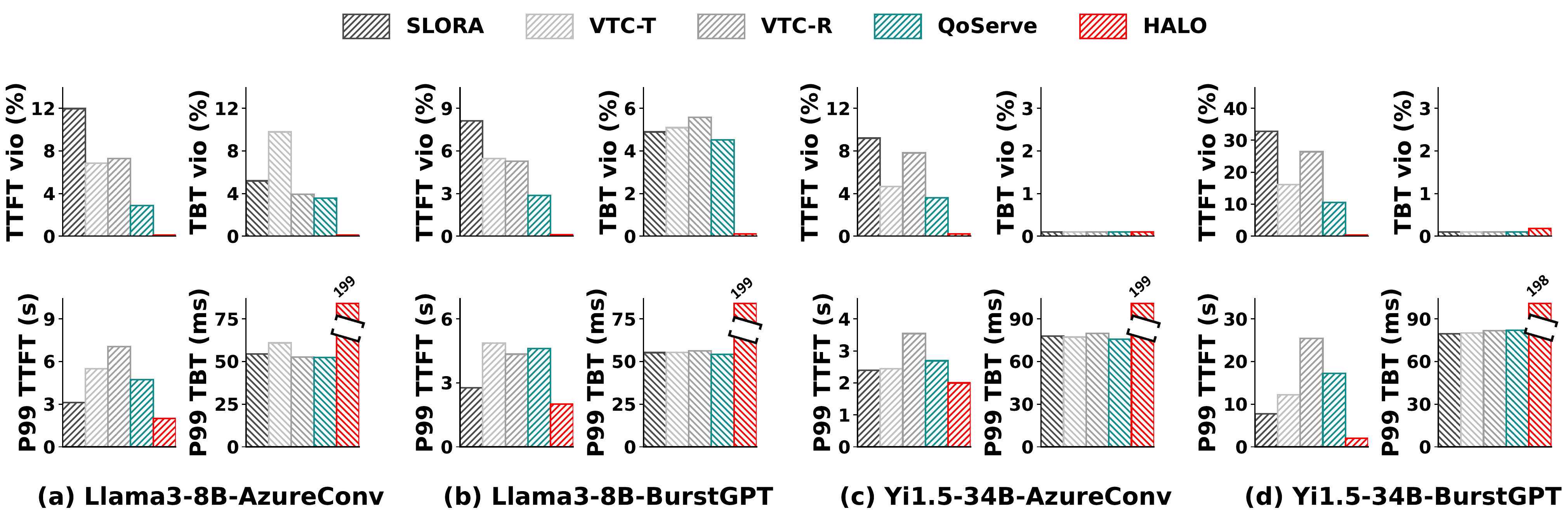}
\vspace{-4mm}
\caption{P99 TTFT and TBT for Llama3-8B and Yi1.5-34B on real-world AzureConv and BurstGPT workloads.
}
\label{fig:exp6-2-1}
\vspace{-2mm}
\end{figure*}

We first evaluate HALO using Llama3-8B and Yi1.5-34B on real-world workload traces. \autoref{fig:exp6-1-1} shows their request rates after scaling down, since they originate from a large cluster. As shown in the figure, AzureConv is relatively stable while BurstGPT exhibits bursty request patterns.

\autoref{fig:exp6-2-1} shows the latency distribution for TTFT and TBT in the {\it Hospital} scenario. HALO improves P99 TTFT (99\%-ile TTFT) by 2.1$\times$, 3.1$\times$, 5.0$\times$, and 3.6$\times$ over SLoRA, VTC-T, VTC-R, and QoServe respectively. HALO reduces TTFT violations by 15.3\%, 8.1\%, 11.6\%, and 4.8\% compared to SLoRA, VTC-T, VTC-R, and QoServe. HALO also reduces TBT violations by 2.5\%, 3.7\%, 2.3\%, and 2.0\% compared to SLoRA, VTC-T, VTC-R, and QoServe.

The performance improvement comes from two key factors. First, HALO accelerates LoRA execution by enabling parallel execution of the base layer and the LoRA layer. In contrast, all baselines suffer from the inefficiencies of Punica, which limits end-to-end latency improvements. By using base-LoRA parallelism, HALO reduces decode latency and directly raises the upper bound on system throughput.

Second, while the baselines can distinguish requests by service type, they cannot exploit SLO differences across requests. HALO explicitly leverages request-level slack and SLO differences to guide scheduling decisions. Its SLO-aware scheduling strategy uses accumulated slack to enable more flexible scheduling. Specifically, HALO returns each token according to its SLO target. This introduces higher P99 TBT but improves both TTFT and TBT guarantees.

Note that, for the baselines, Llama3-8B and Qwen2.5-7B suffer from severe TBT violations, while Yi1.5-34B and Qwen2.5-32B do not. This is because the maximum batch size for Llama3-8B and Qwen2.5-7B is set to 384. Their decode time at 1K sequence length is 45 to 46ms. When the sequence length increases further, they easily suffer TBT violations. The maximum batch size for Yi1.5-34B and Qwen2.5-32B is set to 256. Their decode time at 1K sequence length is 70 to 73ms. Even when the sequence length increases further, they do not suffer TBT violations. In contrast, HALO exploits TBT slack to coordinate the computation order. This allows it to avoid all TBT violations across every model.

\begin{figure*}
\centering
\includegraphics[width=0.8\columnwidth]{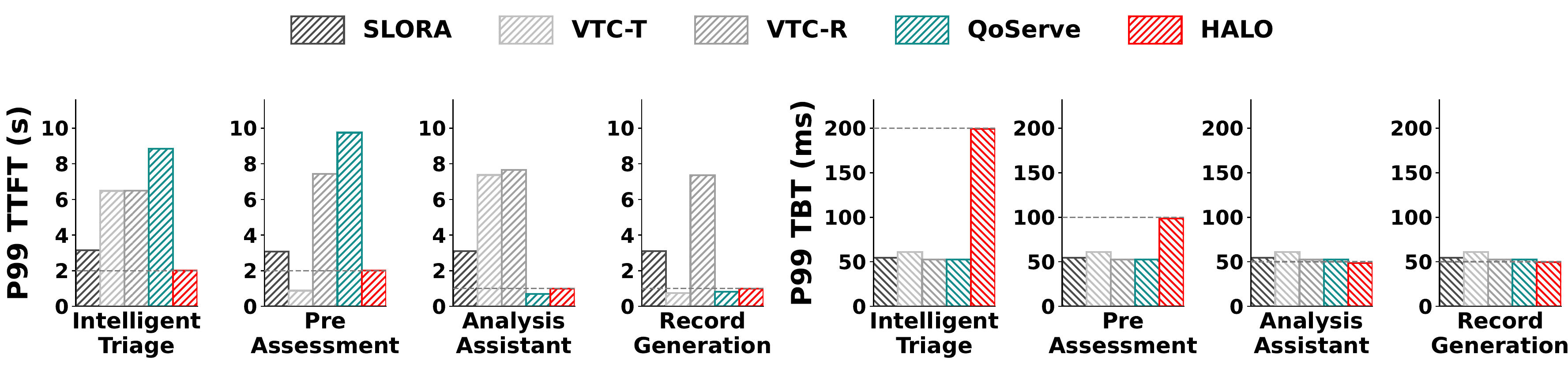}
\vspace{-4mm}
\caption{P99 TTFT and TBT for Llama3-8B on the AzureConv workload with all methods.}
\vspace{-4mm}
\label{fig:exp6-2-2}
\end{figure*}

\autoref{fig:exp6-2-2} shows the P99 TTFT and P99 TBT for four services. As shown, the two services with strict SLOs (analysis and record) suffer severe TBT violations, while the two services with relaxed SLOs (triage and preconsult) have their TBT targets fully met. This further demonstrates that prior works cannot exploit TBT slack. Also, it is worth noting that the P99 TBT for the two strict SLO services is 70ms against a TBT target of 50ms. This is because the TBT target is set based on a 1K context length, while actual requests may have a 4K context length. If we set a smaller chunk size based on 4K context length, TTFT violations will increase dramatically. In contrast, since HALO performs SLO-aware scheduling, it avoids TBT violations caused by varying context lengths.

For VTC-T and VTC-R, they can only control the order in which requests enter the decode batch. When VTC-T and VTC-R adjust the request serving order based on TTFT violations, they can achieve a fair TTFT violation rate across the four services. However, they cannot resolve TTFT violations and TBT violations, since they are unaware of TBT target differences across requests.

For QoServe, it uses the earliest deadline first strategy to adjust the order in which requests enter the batch. This reduces TTFT violations to some extent. However, it also cannot exploit the SLO slack from TBT differences. Although QoServe can dynamically adjust the chunk size based on the strictest TBT target in the running batch, all four services in this scenario are user-facing services. There is little opportunity for QoServe to benefit from adjusting the chunk size.

\subsection{Evaluation in Service Center Scenario}

\begin{figure*}
\centering
\includegraphics[width=\columnwidth]{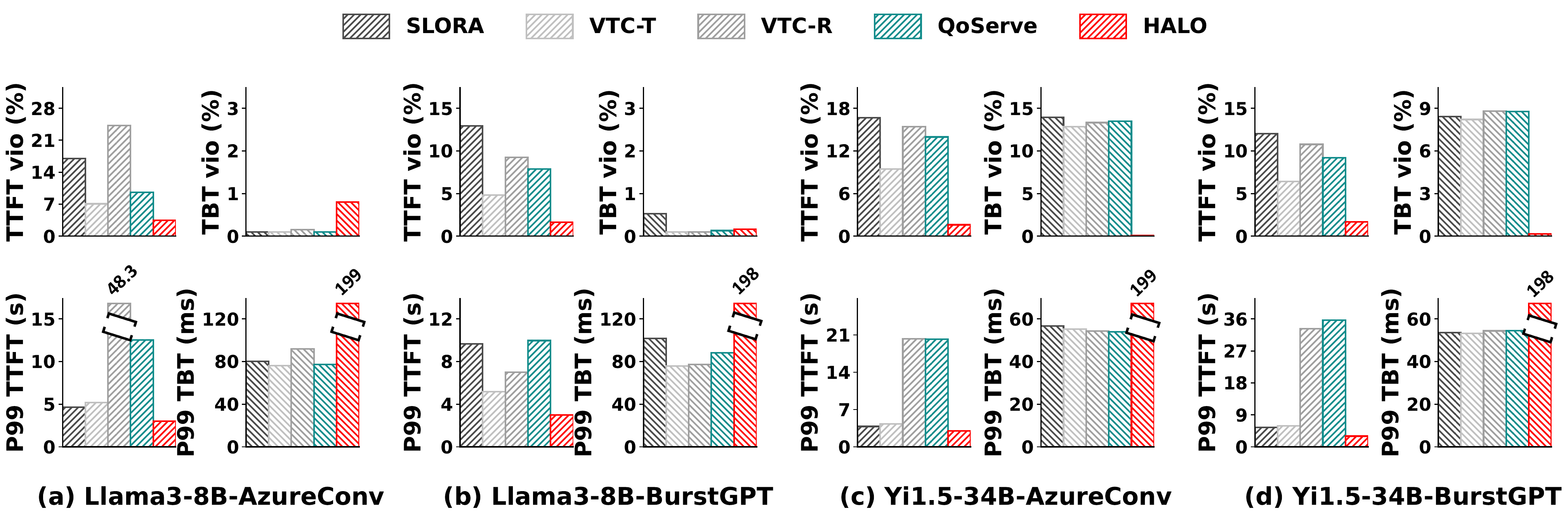}
\caption{P99 TTFT and TBT for Llama3-8B and Yi1.5-34B on real-world AzureConv and BurstGPT workloads.
}
\label{fig:exp6-3-1}
\end{figure*}

\autoref{fig:exp6-3-1} shows the latency distribution for TTFT and TBT in the {\it Service Center} Scenario. HALO improves P99 TTFT (99\%-ile TTFT) by 1.9$\times$, 1.7$\times$, 9.1$\times$, and 6.5$\times$ over SLoRA, VTC-T, VTC-R, and QoServe respectively. HALO reduces TTFT violations by 12.6\%, 4.8\%, 12.8\%, and 8.1\% compared to SLoRA, VTC-T, VTC-R, and QoServe. HALO also reduces TBT violations by 5.4\%, 5.0\%, 5.2\%, and 5.3\% compared to SLoRA, VTC-T, VTC-R, and QoServe.

Although the hospital and service center scenarios involve different LLM services, the results above show that HALO consistently achieves strong performance gains in both settings. This demonstrates that HALO can effectively support diverse on-premise LLM deployment scenarios, even when they differ in TTFT and TBT targets, request characteristics, and underlying datasets.

\subsection{SLO Attainment with Different RPS}

We also evaluate the SLO attainment of different systems under varying RPS. Following prior works \cite{chunked, splitwise, mooncake, slora}, the request arrival pattern at each RPS follows a Poisson distribution. Four types of requests are generated with equal probability using the corresponding dataset and SLO targets from the hospital scenario. We stop testing once the serving system becomes unstable or fails to meet the TTFT target.

\begin{figure*}
\centering
\includegraphics[width=\columnwidth]{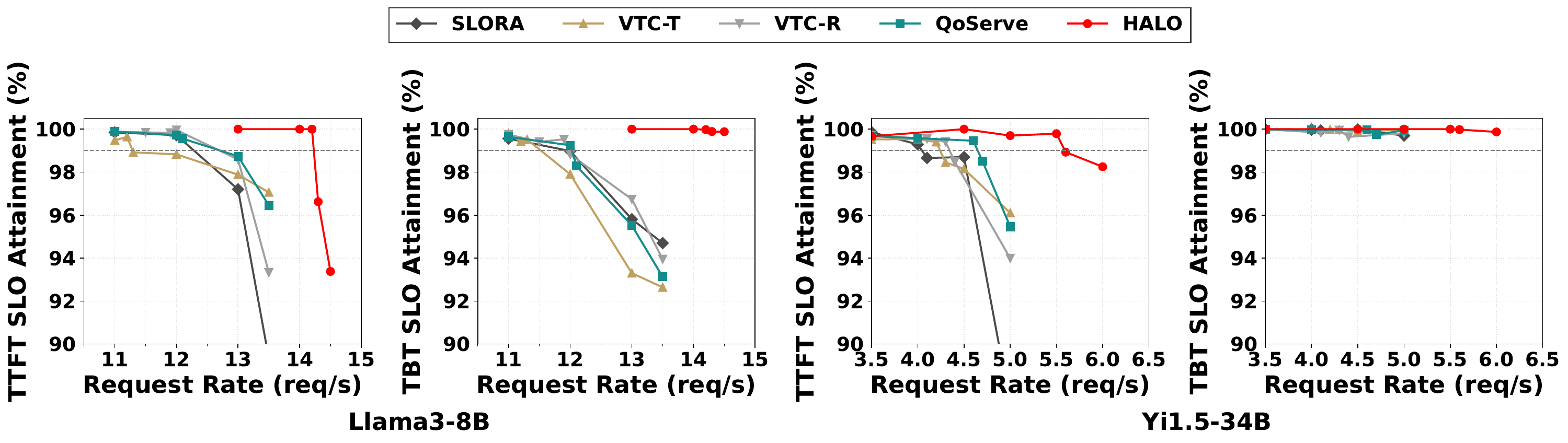}
\caption{SLO Attainment with Different RPS for Llama3-8B and Yi1.5-34B in hospital Scenario.
}
\label{fig:exp6-4}
\end{figure*}

\autoref{fig:exp6-4} shows the SLO attainment results under gradually increasing workloads. Under the constraint of meeting P99 TTFT guarantees, HALO achieves 29.2\%, 29.5\%, 24.8\%, and 19.5\% throughput improvement compared to SLoRA, VTC-T, VTC-R, and QoServe. HALO still provides P99 TTFT and P99 TBT guarantees at higher RPS. This again comes from two factors: base-LoRA parallelism and SLO-aware scheduling.

Specifically, as shown in \autoref{fig:exp6-4}, all baselines begin to suffer TTFT violations once RPS exceeds 13. This is because they cannot exploit TBT slack to enable earlier prefill computation. Meantime, all baselines also suffer some TBT violations, even though chunked-prefill sets the maximum batch size to prioritize decode computation. This is because long context requests (like 4K length) cause some decode iterations to exceed the TBT target (set based on 1K length). In contrast, HALO exploits TBT slack to serve newly arrived requests faster and reduce TTFT violations. It also uses TBT slack to mitigate TBT violations caused by long context lengths.

\subsection{Different LLM models}

\begin{figure*}
\centering
\includegraphics[width=\columnwidth]{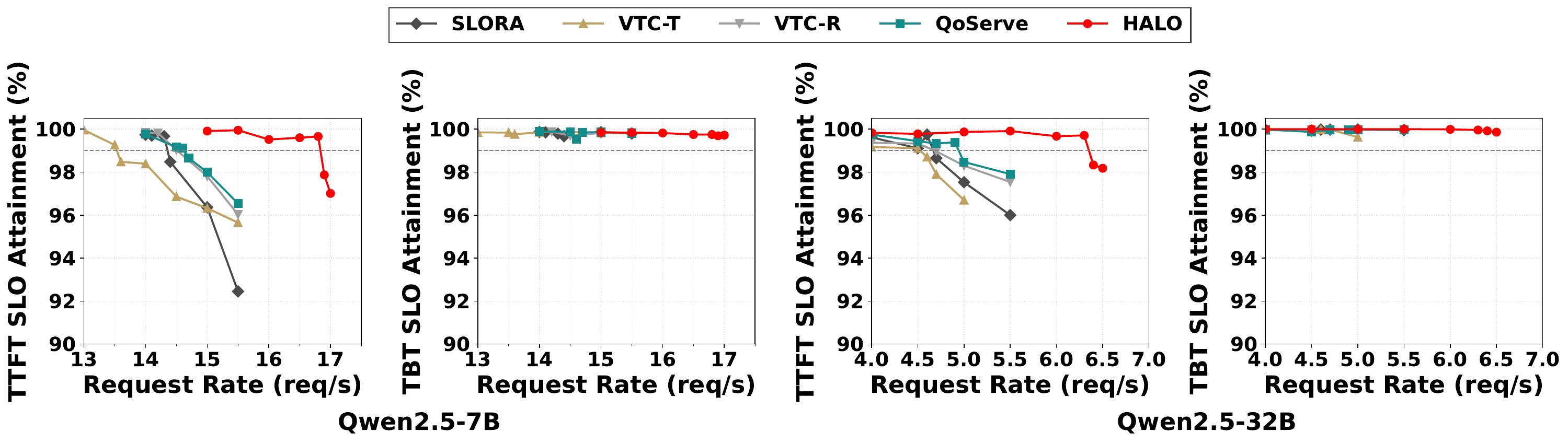}
\caption{SLO Attainment with Different RPS for Qwen2.5-7B and Qwen2.5-32B in hospital Scenario.
}
\label{fig:exp6-5}
\end{figure*}

Beyond the Llama series, we also select models from the Qwen series for evaluation. Specifically, we choose Qwen2.5-7B and Qwen2.5-32B. \autoref{fig:exp6-5} shows the SLO attainment results under gradually increasing workloads. Under the constraint of meeting P99 TTFT guarantees, HALO achieves 27.2\%, 32.2\%, 25.0\%, and 21.8\% throughput improvement compared to SLoRA, VTC-T, VTC-R, and QoServe. These results show that HALO also achieves significant performance improvement on new models. This is because all designs in HALO are model-agnostic. Its designs are rooted in base-LoRA layer parallelism and SLO-aware scheduling, which are independent of model architecture.

\subsection{The Effect of Base-LoRA Parallelism}
\label{sec:base-lora}

\begin{figure}
\centering
\includegraphics[width=0.8\linewidth]{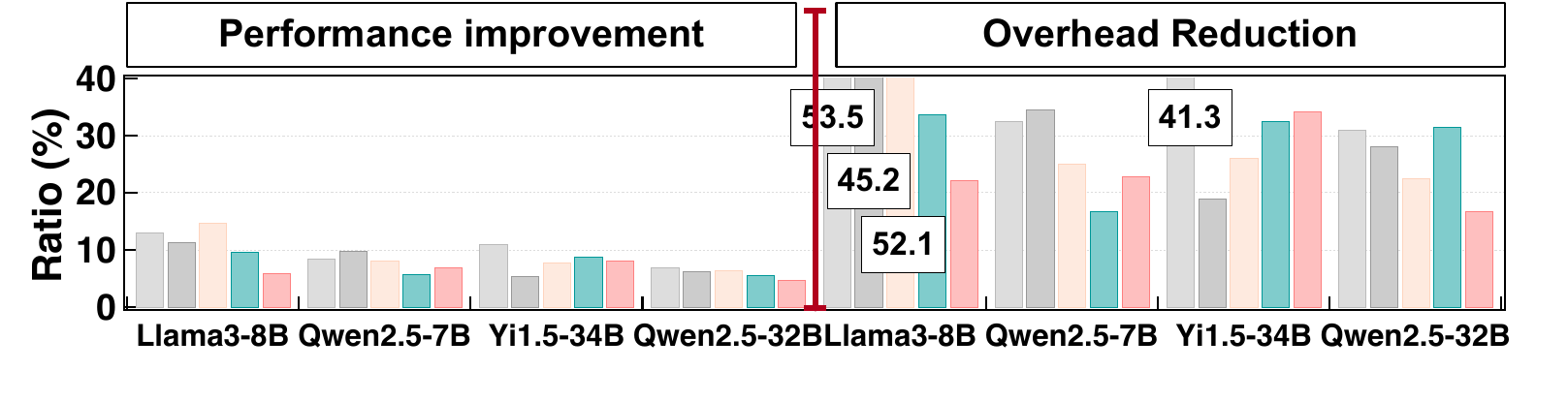}
\caption{The performance improvement and overhead reduction brought the base-lora parallelism.}
\label{fig:exp7-5}
\end{figure}

\autoref{fig:exp7-5} shows the end-to-end decode latency improvement and the reduction in LoRA-induced overhead for four models under different batch sizes. As shown in the figure, when the sequence length is 128, base-LoRA parallelism achieves an average performance improvement of 8.3\%, with a maximum improvement of 14.7\%. It reduces LoRA-induced execution overhead by 31.1\% on average and by up to 53.5\% at maximum. These results demonstrate the effectiveness of base LoRA parallelism.

When the sequence length increases to 1024, the average performance improvement decreases to 6.4\%. This reduction occurs because, although base-LoRA parallelism itself is insensitive to sequence length, another major component of inference, attention computation, increases with sequence length. As the baseline decode latency grows, the relative benefit of base-LoRA parallelism becomes smaller. We also note that prefill is computationally intensive, and base-LoRA parallelism does not improve performance during the prefill phase.

\subsection{Ablation Study}

In this experiment, we conduct ablation studies to understand the contribution of each component of HALO. We construct HALO-P, which enables only base-LoRA parallelism, and HALO-S, which enables only SLO-aware scheduling. The experiment is performed on Llama3-8B using a single A100 GPU.

\autoref{fig:exp7-6}(a) shows the latency distribution for TTFT and TBT in the Hospital Scenario. HALO achieves 18.3\%, 4.1\%, and 10.3\% throughput improvement compared to SLoRA, HALO-P, and HALO-S, respectively.
The results show that enabling only base-LoRA parallelism still leads to significant SLO violations. In contrast, enabling only SLO-aware scheduling is constrained by sequential base-LoRA execution. These findings indicate that both components are necessary to achieve the performance benefits of HALO.

\subsection{Different GPU}

In this experiment, we evaluate HALO on a different GPU to verify its effectiveness across hardware platforms. Specifically, we run Llama3-8B on a single Ada6000 GPU.

\begin{figure}
\centering
\includegraphics[width=\columnwidth]{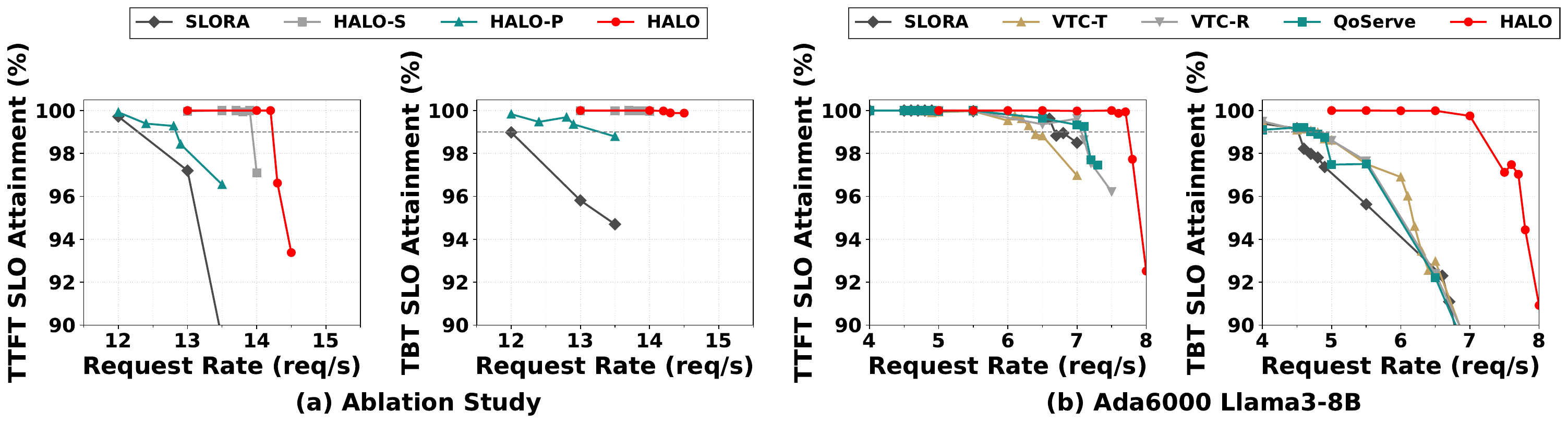}
\caption{The throughput and violation in hospital Scenario:
\textbf{(a)} {\it SLORA}, HALO-P, HALO-S and \sysname{};
\textbf{(b)} {\it SLORA}, {\it VTC-T}, {\it VTC-R}, {\it QoServe} and \sysname{} for {\it Llama3-8B} on the {\it Ada6000}.}
\label{fig:exp7-6}
\end{figure}

\autoref{fig:exp7-6}(b) shows the TTFT and TBT latency distributions in the hospital scenario. Under the constraint of meeting P99 TTFT guarantees, HALO achieves 23.0\%, 27.4\%, 19.9\%, and 19.5\% throughput improvement compared to SLORA, VTC-T, VTC-R, and QoServe. The throughput improvement comes from both TTFT and TBT optimization. As shown in the figure, all baselines suffer from severe TBT violations. This is because Llama3-8B with a batch size of 256 at sequence length 256 requires 49ms for decode computation. When the input length exceeds 256, the baselines suffer from severe TBT violations. On the other hand, if we set the maximum batch size below 256 such as 128, the decode computation encounters severe GPU underutilization. The results show that HALO continues to deliver strong performance and efficiently support GPUs beyond the A100.

\subsection{Overhead}

HALO's overhead primarily comes from two sources. Firstly, HALO needs to determine the base-LoRA partition during the serving startup phase. Secondly, during runtime scheduling, HALO needs to decide which requests enter the running batch.

For the partition search, the process requires 12.4 profiling runs on average. For a typical setup ($N_{lora}=4, N_{bs}=8, N_{conf}=12.4, T_{decode} = 100ms$), the profiling time exceeds 39.7 seconds per model, which is tolerable for the framework startup. For the runtime scheduling, the sorting process on the CPU side is implemented using an asynchronous way, which involves no scheduling overhead.

%% file: contents/related.tex
\section{Related works}

\paragraph{LLM serving optimization:} LLM serving optimization has attracted great attention. Existing work explores several directions, including compute management\cite{splitwise, chunked, zhong2024distserve}, memory management\cite{kwon2023efficient, zheng2023efficiently, lin2024parrot}, and GPU kernel optimization\cite{dao2022flashattention, ye2025flashinfer}. Splitwise\cite{splitwise} disaggregates LLM serving into separate prefill and decode instances, while chunked prefill\cite{chunked} divides the prefill phase into smaller chunks and fuses each chunk with a decode iteration. PagedAttention\cite{kwon2023efficient} introduces a paged memory pool that enables reuse of key value caches across the prefill and decode phases. SGLang\cite{zheng2023efficiently} exploits context aware caching to maximize reuse of key value segments across requests. FlashAttention\cite{dao2022flashattention} and FlashInfer\cite{ye2025flashinfer} focus on designing specialized kernels to accelerate attention computation. These approaches target general improvements to the serving framework, and are orthogonal to the techniques proposed in HALO.

\paragraph{LoRA serving optimization:} Existing works\cite{slora, lorax, li2024caraserve, punica, wu2024dlora, elora} on LoRA serving optimization largely fall into two categories. One line of work targets fast switching among multiple LoRAs, while the other focuses on reducing LoRA execution latency. SLoRA\cite{slora} enables scalable serving of thousands of adapters by introducing unified paging. LoRAX\cite{lorax} supports just in time dynamic loading of adapters through a tiered caching (GPU, CPU, and disk). CaraServe\cite{li2024caraserve} concentrates on the synchronization between CPU and GPU to hide the latency of fetching adapters. These systems are designed to handle scenarios where thousands of LoRA adapters. In contrast, an on-premise deployment typically supports only a small number of LoRA adapters, which makes these approaches ineffective. In addition, Punica\cite{punica} proposes a new CUDA kernel to accelerate LoRA execution. However, our results show that this optimization is still far from optimal when considering end-to-end latency.

\paragraph{LLM scheduling strategy:} Recent researches\cite{loraserve, fastserve, vtc, sloserve} have devoted substantial effort to LLM scheduling strategies that aim to minimize average job completion time. LoRAServe\cite{loraserve} applies dynamic adapter placement and request routing to balance load across multiple GPUs. FastServe\cite{fastserve} uses preemptive scheduling to mitigate head of line blocking caused by long running requests. VTC\cite{vtc} introduces a deficit round robin scheduler based on virtual tokens to provide fairness in LLM serving. SLOs-Serve\cite{sloserve} supports multiple SLOs and multi-instance deployment through dynamic request routing. While these approaches can reduce job completion time in certain scenarios, they do not address multi-SLO optimization in the on-premise LLM deployment. This is because the on-premise deployment generally has a single instance and tightly constrained computational resources.

%% file: contents/conclusion.tex
\section{Conclusion\label{sec:conclusion}}

In this paper, we present \sysname{}, a specialized scheduling framework that addresses the performance bottlenecks of LoRA-enabled on-premise LLM servers in privacy-sensitive sectors. By introducing a spatial multiplexing strategy that partitions GPU Streaming Multiprocessors to overlap Base and LoRA computations, and an SLO-aware scheduler driven by "request-level slack," we successfully decouple execution from the constraints of rigid batching. Our evaluation demonstrates that \sysname{} effectively mitigates resource contention and significantly reduces SLO violations while improving aggregate throughput compared to state-of-the-art baselines, providing a robust foundation for secure, high-performance local LLM inference.

%% file: main.bbl

\begin{thebibliography}{43}


\ifx \showCODEN    \undefined \def \showCODEN     #1{\unskip}     \fi
\ifx \showISBNx    \undefined \def \showISBNx     #1{\unskip}     \fi
\ifx \showISBNxiii \undefined \def \showISBNxiii  #1{\unskip}     \fi
\ifx \showISSN     \undefined \def \showISSN      #1{\unskip}     \fi
\ifx \showLCCN     \undefined \def \showLCCN      #1{\unskip}     \fi
\ifx \shownote     \undefined \def \shownote      #1{#1}          \fi
\ifx \showarticletitle \undefined \def \showarticletitle #1{#1}   \fi
\ifx \showURL      \undefined \def \showURL       {\relax}        \fi
\providecommand\bibfield[2]{#2}
\providecommand\bibinfo[2]{#2}
\providecommand\natexlab[1]{#1}
\providecommand\showeprint[2][]{arXiv:#2}
\makeatletter
\@ifundefined{NAT@parse@date}{}{\let\NAT@parse@date@orig\NAT@parse@date}
\@ifundefined{NAT@parse@date}{}{\def\NAT@parse@date#1#2#3#4#5#6@@{\NAT@parse@date@orig#1#2#3#4#5#6@@\def\NAT@tempyear{0000}\def\NAT@tempexlab{{?}}\ifx\NAT@year\NAT@tempyear\ifx\NAT@exlab\NAT@tempexlab\def\NAT@date{[n.\,d.]}\else\edef\NAT@date{[n.\,d.]\NAT@exlab}\fi\fi}}
\makeatother

\bibitem[Agrawal et~al\mbox{.}(2024)]%
        {chunked}
\bibfield{author}{\bibinfo{person}{Amey Agrawal}, \bibinfo{person}{Nitin Kedia}, \bibinfo{person}{Ashish Panwar}, \bibinfo{person}{Jayashree Mohan}, \bibinfo{person}{Nipun Kwatra}, \bibinfo{person}{Bhargav Gulavani}, \bibinfo{person}{Alexey Tumanov}, {and} \bibinfo{person}{Ramachandran Ramjee}.} \bibinfo{year}{2024}\natexlab{}.
\newblock \showarticletitle{Taming $\{$Throughput-Latency$\}$ tradeoff in $\{$LLM$\}$ inference with $\{$Sarathi-Serve$\}$}. In \bibinfo{booktitle}{\emph{18th USENIX Symposium on Operating Systems Design and Implementation (OSDI 24)}}. \bibinfo{pages}{117--134}.
\newblock


\bibitem[AI et~al\mbox{.}(2024)]%
        {ai2024yi}
\bibfield{author}{\bibinfo{person}{01. AI}, \bibinfo{person}{:}, \bibinfo{person}{Alex Young}, \bibinfo{person}{Bei Chen}, \bibinfo{person}{Chao Li}, \bibinfo{person}{Chengen Huang}, \bibinfo{person}{Ge Zhang}, \bibinfo{person}{Guanwei Zhang}, \bibinfo{person}{Heng Li}, \bibinfo{person}{Jiangcheng Zhu}, \bibinfo{person}{Jianqun Chen}, \bibinfo{person}{Jing Chang}, \bibinfo{person}{Kaidong Yu}, \bibinfo{person}{Peng Liu}, \bibinfo{person}{Qiang Liu}, \bibinfo{person}{Shawn Yue}, \bibinfo{person}{Senbin Yang}, \bibinfo{person}{Shiming Yang}, \bibinfo{person}{Tao Yu}, \bibinfo{person}{Wen Xie}, \bibinfo{person}{Wenhao Huang}, \bibinfo{person}{Xiaohui Hu}, \bibinfo{person}{Xiaoyi Ren}, \bibinfo{person}{Xinyao Niu}, \bibinfo{person}{Pengcheng Nie}, \bibinfo{person}{Yuchi Xu}, \bibinfo{person}{Yudong Liu}, \bibinfo{person}{Yue Wang}, \bibinfo{person}{Yuxuan Cai}, \bibinfo{person}{Zhenyu Gu}, \bibinfo{person}{Zhiyuan Liu}, {and} \bibinfo{person}{Zonghong Dai}.} \bibinfo{year}{2024}\natexlab{}.
\newblock \bibinfo{title}{Yi: Open Foundation Models by 01.AI}.
\newblock
\showeprint[arxiv]{2403.04652}~[cs.CL]


\bibitem[anon8231489123(0000)]%
        {sharegpt}
\bibfield{author}{\bibinfo{person}{anon8231489123}.} \bibinfo{year}{[n.\,d.]}\natexlab{}.
\newblock \bibinfo{title}{Dataset-ShareGPT}.
\newblock \bibinfo{howpublished}{\url{https://huggingface.co/datasets/anon8231489123/ShareGPT_Vicuna_unfiltered}}.
\newblock
\newblock
\shownote{Accessed: 2026-01-27}.


\bibitem[Bojar et~al\mbox{.}(2014)]%
        {wmt}
\bibfield{author}{\bibinfo{person}{Ondrej Bojar}, \bibinfo{person}{Christian Buck}, \bibinfo{person}{Christian Federmann}, \bibinfo{person}{Barry Haddow}, \bibinfo{person}{Philipp Koehn}, \bibinfo{person}{Johannes Leveling}, \bibinfo{person}{Christof Monz}, \bibinfo{person}{Pavel Pecina}, \bibinfo{person}{Matt Post}, \bibinfo{person}{Herve Saint-Amand}, \bibinfo{person}{Radu Soricut}, \bibinfo{person}{Lucia Specia}, {and} \bibinfo{person}{Ale~{s} Tamchyna}.} \bibinfo{year}{2014}\natexlab{}.
\newblock \showarticletitle{Findings of the 2014 Workshop on Statistical Machine Translation}. In \bibinfo{booktitle}{\emph{Proceedings of the Ninth Workshop on Statistical Machine Translation}}. \bibinfo{publisher}{Association for Computational Linguistics}, \bibinfo{address}{Baltimore, Maryland, USA}, \bibinfo{pages}{12--58}.
\newblock
\urldef\tempurl%
\url{http://www.aclweb.org/anthology/W/W14/W14-3302}
\showURL{%
\tempurl}


\bibitem[Burns et~al\mbox{.}(2025)]%
        {llmcost}
\bibfield{author}{\bibinfo{person}{Michael~L Burns}, \bibinfo{person}{Ssu-Ying Chen}, \bibinfo{person}{Chu-An Tsai}, \bibinfo{person}{John Vandervest}, \bibinfo{person}{Balaji Pandian}, \bibinfo{person}{Paige Nong}, \bibinfo{person}{David~A Hanauer}, \bibinfo{person}{Andrew Rosenberg}, {and} \bibinfo{person}{Jodyn Platt}.} \bibinfo{year}{2025}\natexlab{}.
\newblock \showarticletitle{Generative AI costs in large healthcare systems, an example in revenue cycle}.
\newblock \bibinfo{journal}{\emph{NPJ digital medicine}} \bibinfo{volume}{8}, \bibinfo{number}{1} (\bibinfo{year}{2025}), \bibinfo{pages}{579}.
\newblock


\bibitem[Chen et~al\mbox{.}(2024)]%
        {punica}
\bibfield{author}{\bibinfo{person}{Lequn Chen}, \bibinfo{person}{Zihao Ye}, \bibinfo{person}{Yongji Wu}, \bibinfo{person}{Danyang Zhuo}, \bibinfo{person}{Luis Ceze}, {and} \bibinfo{person}{Arvind Krishnamurthy}.} \bibinfo{year}{2024}\natexlab{}.
\newblock \showarticletitle{Punica: Multi-tenant lora serving}.
\newblock \bibinfo{journal}{\emph{Proceedings of Machine Learning and Systems}}  \bibinfo{volume}{6} (\bibinfo{year}{2024}), \bibinfo{pages}{1--13}.
\newblock


\bibitem[Chen et~al\mbox{.}(2025)]%
        {sloserve}
\bibfield{author}{\bibinfo{person}{Siyuan Chen}, \bibinfo{person}{Zhipeng Jia}, \bibinfo{person}{Samira Khan}, \bibinfo{person}{Arvind Krishnamurthy}, {and} \bibinfo{person}{Phillip~B Gibbons}.} \bibinfo{year}{2025}\natexlab{}.
\newblock \showarticletitle{SLOs-Serve: Optimized Serving of Multi-SLO LLMs}.
\newblock \bibinfo{journal}{\emph{arXiv preprint arXiv:2504.08784}} (\bibinfo{year}{2025}).
\newblock


\bibitem[Dao et~al\mbox{.}(2022)]%
        {dao2022flashattention}
\bibfield{author}{\bibinfo{person}{Tri Dao}, \bibinfo{person}{Dan Fu}, \bibinfo{person}{Stefano Ermon}, \bibinfo{person}{Atri Rudra}, {and} \bibinfo{person}{Christopher R{\'e}}.} \bibinfo{year}{2022}\natexlab{}.
\newblock \showarticletitle{Flashattention: Fast and memory-efficient exact attention with io-awareness}.
\newblock \bibinfo{journal}{\emph{Advances in neural information processing systems}}  \bibinfo{volume}{35} (\bibinfo{year}{2022}), \bibinfo{pages}{16344--16359}.
\newblock


\bibitem[DeepSeek-AI(2025)]%
        {deepseekai2025deepseekr1incentivizingreasoningcapability}
\bibfield{author}{\bibinfo{person}{DeepSeek-AI}.} \bibinfo{year}{2025}\natexlab{}.
\newblock \bibinfo{title}{DeepSeek-R1: Incentivizing Reasoning Capability in LLMs via Reinforcement Learning}.
\newblock
\showeprint[arxiv]{2501.12948}~[cs.CL]
\urldef\tempurl%
\url{https://arxiv.org/abs/2501.12948}
\showURL{%
\tempurl}


\bibitem[DELL(0000)]%
        {dell}
\bibfield{author}{\bibinfo{person}{DELL}.} \bibinfo{year}{[n.\,d.]}\natexlab{}.
\newblock \bibinfo{title}{The Dell AI Factory with NVIDIA: Accelerating AI Adoption for the Enterprise.}
\newblock \bibinfo{howpublished}{\url{https://www.nvidia.com/en-us/on-demand/session/aisummitin24-siin1250/}}.
\newblock
\newblock
\shownote{Accessed: 2026-01-27}.


\bibitem[Friha et~al\mbox{.}(2024)]%
        {friha2024llm}
\bibfield{author}{\bibinfo{person}{Othmane Friha}, \bibinfo{person}{Mohamed~Amine Ferrag}, \bibinfo{person}{Burak Kantarci}, \bibinfo{person}{Burak Cakmak}, \bibinfo{person}{Arda Ozgun}, {and} \bibinfo{person}{Nassira Ghoualmi-Zine}.} \bibinfo{year}{2024}\natexlab{}.
\newblock \showarticletitle{Llm-based edge intelligence: A comprehensive survey on architectures, applications, security and trustworthiness}.
\newblock \bibinfo{journal}{\emph{IEEE Open Journal of the Communications Society}} (\bibinfo{year}{2024}).
\newblock


\bibitem[Goel et~al\mbox{.}(2026)]%
        {qoserve}
\bibfield{author}{\bibinfo{person}{Kanishk Goel}, \bibinfo{person}{Jayashree Mohan}, \bibinfo{person}{Nipun Kwatra}, \bibinfo{person}{Ravi~Shreyas Anupindi}, {and} \bibinfo{person}{Ramachandran Ramjee}.} \bibinfo{year}{2026}\natexlab{}.
\newblock \showarticletitle{QoServe: Breaking the Silos of LLM Inference Serving}. In \bibinfo{booktitle}{\emph{Proceedings of the 31st ACM International Conference on Architectural Support for Programming Languages and Operating Systems, Volume 2}}. \bibinfo{pages}{1492--1507}.
\newblock


\bibitem[Grattafiori et~al\mbox{.}(2024)]%
        {llama3}
\bibfield{author}{\bibinfo{person}{Aaron Grattafiori}, \bibinfo{person}{Abhimanyu Dubey}, \bibinfo{person}{Abhinav Jauhri}, \bibinfo{person}{Abhinav Pandey}, \bibinfo{person}{Abhishek Kadian}, \bibinfo{person}{Ahmad Al-Dahle}, \bibinfo{person}{Aiesha Letman}, \bibinfo{person}{Akhil Mathur}, \bibinfo{person}{Alan Schelten}, \bibinfo{person}{Alex Vaughan}, {et~al\mbox{.}}} \bibinfo{year}{2024}\natexlab{}.
\newblock \showarticletitle{The llama 3 herd of models}.
\newblock \bibinfo{journal}{\emph{arXiv preprint arXiv:2407.21783}} (\bibinfo{year}{2024}).
\newblock


\bibitem[Hu et~al\mbox{.}(2022)]%
        {hu2022lora}
\bibfield{author}{\bibinfo{person}{Edward~J Hu}, \bibinfo{person}{Yelong Shen}, \bibinfo{person}{Phillip Wallis}, \bibinfo{person}{Zeyuan Allen-Zhu}, \bibinfo{person}{Yuanzhi Li}, \bibinfo{person}{Shean Wang}, \bibinfo{person}{Lu Wang}, \bibinfo{person}{Weizhu Chen}, {et~al\mbox{.}}} \bibinfo{year}{2022}\natexlab{}.
\newblock \showarticletitle{Lora: Low-rank adaptation of large language models.}
\newblock \bibinfo{journal}{\emph{ICLR}} \bibinfo{volume}{1}, \bibinfo{number}{2} (\bibinfo{year}{2022}), \bibinfo{pages}{3}.
\newblock


\bibitem[Jaiswal et~al\mbox{.}(2025)]%
        {loraserve}
\bibfield{author}{\bibinfo{person}{Shashwat Jaiswal}, \bibinfo{person}{Shrikara Arun}, \bibinfo{person}{Anjaly Parayil}, \bibinfo{person}{Ankur Mallick}, \bibinfo{person}{Spyros Mastorakis}, \bibinfo{person}{Alind Khare}, \bibinfo{person}{Chloi Alverti}, \bibinfo{person}{Renee~St Amant}, \bibinfo{person}{Chetan Bansal}, \bibinfo{person}{Victor R{\"u}hle}, {et~al\mbox{.}}} \bibinfo{year}{2025}\natexlab{}.
\newblock \showarticletitle{Serving Heterogeneous LoRA Adapters in Distributed LLM Inference Systems}.
\newblock \bibinfo{journal}{\emph{arXiv preprint arXiv:2511.22880}} (\bibinfo{year}{2025}).
\newblock


\bibitem[Kwon et~al\mbox{.}(2023)]%
        {kwon2023efficient}
\bibfield{author}{\bibinfo{person}{Woosuk Kwon}, \bibinfo{person}{Zhuohan Li}, \bibinfo{person}{Siyuan Zhuang}, \bibinfo{person}{Ying Sheng}, \bibinfo{person}{Lianmin Zheng}, \bibinfo{person}{Cody~Hao Yu}, \bibinfo{person}{Joseph Gonzalez}, \bibinfo{person}{Hao Zhang}, {and} \bibinfo{person}{Ion Stoica}.} \bibinfo{year}{2023}\natexlab{}.
\newblock \showarticletitle{Efficient memory management for large language model serving with pagedattention}. In \bibinfo{booktitle}{\emph{Proceedings of the 29th symposium on operating systems principles}}. \bibinfo{pages}{611--626}.
\newblock


\bibitem[lavita(0000)]%
        {chatdoctor}
\bibfield{author}{\bibinfo{person}{lavita}.} \bibinfo{year}{[n.\,d.]}\natexlab{}.
\newblock \bibinfo{title}{Dataset-ChatDoctor}.
\newblock \bibinfo{howpublished}{\url{https://huggingface.co/datasets/lavita/ChatDoctor-HealthCareMagic-100k}}.
\newblock
\newblock
\shownote{Accessed: 2026-01-27}.


\bibitem[Li et~al\mbox{.}(2025b)]%
        {mobilora}
\bibfield{author}{\bibinfo{person}{Borui Li}, \bibinfo{person}{Yitao Wang}, \bibinfo{person}{Haoran Ma}, \bibinfo{person}{Ligeng Chen}, \bibinfo{person}{Jun Xiao}, {and} \bibinfo{person}{Shuai Wang}.} \bibinfo{year}{2025}\natexlab{b}.
\newblock \showarticletitle{MobiLoRA: Accelerating LoRA-Based LLM Inference on Mobile Devices via Context-Aware KV Cache Optimization}. In \bibinfo{booktitle}{\emph{Proceedings of the 63rd Annual Meeting of the Association for Computational Linguistics (Volume 1: Long Papers)}}. \bibinfo{pages}{23400--23410}.
\newblock


\bibitem[Li et~al\mbox{.}(2024)]%
        {li2024caraserve}
\bibfield{author}{\bibinfo{person}{Suyi Li}, \bibinfo{person}{Hanfeng Lu}, \bibinfo{person}{Tianyuan Wu}, \bibinfo{person}{Minchen Yu}, \bibinfo{person}{Qizhen Weng}, \bibinfo{person}{Xusheng Chen}, \bibinfo{person}{Yizhou Shan}, \bibinfo{person}{Binhang Yuan}, {and} \bibinfo{person}{Wei Wang}.} \bibinfo{year}{2024}\natexlab{}.
\newblock \showarticletitle{Caraserve: Cpu-assisted and rank-aware lora serving for generative llm inference}.
\newblock \bibinfo{journal}{\emph{arXiv preprint arXiv:2401.11240}} (\bibinfo{year}{2024}).
\newblock


\bibitem[Li et~al\mbox{.}(2025a)]%
        {adaserve}
\bibfield{author}{\bibinfo{person}{Zikun Li}, \bibinfo{person}{Zhuofu Chen}, \bibinfo{person}{Remi Delacourt}, \bibinfo{person}{Gabriele Oliaro}, \bibinfo{person}{Zeyu Wang}, \bibinfo{person}{Qinghan Chen}, \bibinfo{person}{Shuhuai Lin}, \bibinfo{person}{April Yang}, \bibinfo{person}{Zhihao Zhang}, \bibinfo{person}{Zhuoming Chen}, {et~al\mbox{.}}} \bibinfo{year}{2025}\natexlab{a}.
\newblock \showarticletitle{AdaServe: Accelerating Multi-SLO LLM Serving with SLO-Customized Speculative Decoding}.
\newblock \bibinfo{journal}{\emph{arXiv preprint arXiv:2501.12162}} (\bibinfo{year}{2025}).
\newblock


\bibitem[Lin et~al\mbox{.}(2024)]%
        {lin2024parrot}
\bibfield{author}{\bibinfo{person}{Chaofan Lin}, \bibinfo{person}{Zhenhua Han}, \bibinfo{person}{Chengruidong Zhang}, \bibinfo{person}{Yuqing Yang}, \bibinfo{person}{Fan Yang}, \bibinfo{person}{Chen Chen}, {and} \bibinfo{person}{Lili Qiu}.} \bibinfo{year}{2024}\natexlab{}.
\newblock \showarticletitle{Parrot: Efficient serving of $\{$LLM-based$\}$ applications with semantic variable}. In \bibinfo{booktitle}{\emph{18th USENIX Symposium on Operating Systems Design and Implementation (OSDI 24)}}. \bibinfo{pages}{929--945}.
\newblock


\bibitem[Malikeh1375(0000)]%
        {medicaldata}
\bibfield{author}{\bibinfo{person}{Malikeh1375}.} \bibinfo{year}{[n.\,d.]}\natexlab{}.
\newblock \bibinfo{title}{Dataset-MedicalData}.
\newblock \bibinfo{howpublished}{\url{https://huggingface.co/datasets/Malikeh1375/medical-question-answering-datasets}}.
\newblock
\newblock
\shownote{Accessed: 2026-01-27}.


\bibitem[Maslej et~al\mbox{.}(2025)]%
        {aireport}
\bibfield{author}{\bibinfo{person}{Nestor Maslej}, \bibinfo{person}{Loredana Fattorini}, \bibinfo{person}{Raymond Perrault}, \bibinfo{person}{Yolanda Gil}, \bibinfo{person}{Vanessa Parli}, \bibinfo{person}{Njenga Kariuki}, \bibinfo{person}{Emily Capstick}, \bibinfo{person}{Anka Reuel}, \bibinfo{person}{Erik Brynjolfsson}, \bibinfo{person}{John Etchemendy}, {et~al\mbox{.}}} \bibinfo{year}{2025}\natexlab{}.
\newblock \showarticletitle{Stanford HAI AI Index Report 2025}.
\newblock \bibinfo{journal}{\emph{arXiv preprint arXiv:2504.07139}} (\bibinfo{year}{2025}).
\newblock


\bibitem[OpenAI(2023)]%
        {openai2023gpt4}
\bibfield{author}{\bibinfo{person}{OpenAI}.} \bibinfo{year}{2023}\natexlab{}.
\newblock \bibinfo{title}{GPT-4 Technical Report}.
\newblock
\urldef\tempurl%
\url{https://openai.com/index/gpt-4-research/}
\showURL{%
\tempurl}


\bibitem[Pan et~al\mbox{.}(2025)]%
        {pan2025cost}
\bibfield{author}{\bibinfo{person}{Guanzhong Pan}, \bibinfo{person}{Vishal Chodnekar}, \bibinfo{person}{Abinas Roy}, {and} \bibinfo{person}{Haibo Wang}.} \bibinfo{year}{2025}\natexlab{}.
\newblock \showarticletitle{A Cost-Benefit Analysis of On-Premise Large Language Model Deployment: Breaking Even with Commercial LLM Services}.
\newblock \bibinfo{journal}{\emph{arXiv preprint arXiv:2509.18101}} (\bibinfo{year}{2025}).
\newblock


\bibitem[Patel et~al\mbox{.}(2024)]%
        {splitwise}
\bibfield{author}{\bibinfo{person}{Pratyush Patel}, \bibinfo{person}{Esha Choukse}, \bibinfo{person}{Chaojie Zhang}, \bibinfo{person}{Aashaka Shah}, \bibinfo{person}{{\'I}{\~n}igo Goiri}, \bibinfo{person}{Saeed Maleki}, {and} \bibinfo{person}{Ricardo Bianchini}.} \bibinfo{year}{2024}\natexlab{}.
\newblock \showarticletitle{Splitwise: Efficient generative llm inference using phase splitting}. In \bibinfo{booktitle}{\emph{2024 ACM/IEEE 51st Annual International Symposium on Computer Architecture (ISCA)}}. IEEE, \bibinfo{pages}{118--132}.
\newblock


\bibitem[Predise(2025)]%
        {lorax}
\bibfield{author}{\bibinfo{person}{Predise}.} \bibinfo{year}{2025}\natexlab{}.
\newblock \bibinfo{title}{LoRAX: Multi-LoRA inference server that scales to 1000s of fine-tuned LLMs}.
\newblock
\urldef\tempurl%
\url{https://github.com/predibase/lorax}
\showURL{%
\tempurl}


\bibitem[Qin et~al\mbox{.}(2025)]%
        {mooncake}
\bibfield{author}{\bibinfo{person}{Ruoyu Qin}, \bibinfo{person}{Zheming Li}, \bibinfo{person}{Weiran He}, \bibinfo{person}{Jialei Cui}, \bibinfo{person}{Feng Ren}, \bibinfo{person}{Mingxing Zhang}, \bibinfo{person}{Yongwei Wu}, \bibinfo{person}{Weimin Zheng}, {and} \bibinfo{person}{Xinran Xu}.} \bibinfo{year}{2025}\natexlab{}.
\newblock \showarticletitle{Mooncake: Trading more storage for less computation—a $\{$KVCache-centric$\}$ architecture for serving $\{$LLM$\}$ chatbot}. In \bibinfo{booktitle}{\emph{23rd USENIX conference on file and storage technologies (FAST 25)}}. \bibinfo{pages}{155--170}.
\newblock


\bibitem[Report(2026)]%
        {idc}
\bibfield{author}{\bibinfo{person}{IDC Report}.} \bibinfo{year}{2026}\natexlab{}.
\newblock \bibinfo{title}{Artificial Intelligence Infrastructure Spending}.
\newblock
\urldef\tempurl%
\url{https://my.idc.com/getdoc.jsp?containerId=prUS53894425}
\showURL{%
\tempurl}


\bibitem[Shen et~al\mbox{.}(2025)]%
        {edgelora}
\bibfield{author}{\bibinfo{person}{Zheyu Shen}, \bibinfo{person}{Yexiao He}, \bibinfo{person}{Ziyao Wang}, \bibinfo{person}{Yuning Zhang}, \bibinfo{person}{Guoheng Sun}, \bibinfo{person}{Wanghao Ye}, {and} \bibinfo{person}{Ang Li}.} \bibinfo{year}{2025}\natexlab{}.
\newblock \showarticletitle{EdgeLoRA: An Efficient Multi-Tenant LLM Serving System on Edge Devices}. In \bibinfo{booktitle}{\emph{Proceedings of the 23rd Annual International Conference on Mobile Systems, Applications and Services}}. \bibinfo{pages}{138--153}.
\newblock


\bibitem[Sheng et~al\mbox{.}(2023)]%
        {slora}
\bibfield{author}{\bibinfo{person}{Ying Sheng}, \bibinfo{person}{Shiyi Cao}, \bibinfo{person}{Dacheng Li}, \bibinfo{person}{Coleman Hooper}, \bibinfo{person}{Nicholas Lee}, \bibinfo{person}{Shuo Yang}, \bibinfo{person}{Christopher Chou}, \bibinfo{person}{Banghua Zhu}, \bibinfo{person}{Lianmin Zheng}, \bibinfo{person}{Kurt Keutzer}, {et~al\mbox{.}}} \bibinfo{year}{2023}\natexlab{}.
\newblock \showarticletitle{S-lora: Serving thousands of concurrent lora adapters}.
\newblock \bibinfo{journal}{\emph{arXiv preprint arXiv:2311.03285}} (\bibinfo{year}{2023}).
\newblock


\bibitem[Sheng et~al\mbox{.}(2024)]%
        {vtc}
\bibfield{author}{\bibinfo{person}{Ying Sheng}, \bibinfo{person}{Shiyi Cao}, \bibinfo{person}{Dacheng Li}, \bibinfo{person}{Banghua Zhu}, \bibinfo{person}{Zhuohan Li}, \bibinfo{person}{Danyang Zhuo}, \bibinfo{person}{Joseph~E Gonzalez}, {and} \bibinfo{person}{Ion Stoica}.} \bibinfo{year}{2024}\natexlab{}.
\newblock \showarticletitle{Fairness in serving large language models}. In \bibinfo{booktitle}{\emph{18th USENIX Symposium on Operating Systems Design and Implementation (OSDI 24)}}. \bibinfo{pages}{965--988}.
\newblock


\bibitem[Taori et~al\mbox{.}(2023)]%
        {alpaca}
\bibfield{author}{\bibinfo{person}{Rohan Taori}, \bibinfo{person}{Ishaan Gulrajani}, \bibinfo{person}{Tianyi Zhang}, \bibinfo{person}{Yann Dubois}, \bibinfo{person}{Xuechen Li}, \bibinfo{person}{Carlos Guestrin}, \bibinfo{person}{Percy Liang}, {and} \bibinfo{person}{Tatsunori~B. Hashimoto}.} \bibinfo{year}{2023}\natexlab{}.
\newblock \bibinfo{title}{Stanford Alpaca: An Instruction-following LLaMA model}.
\newblock \bibinfo{howpublished}{\url{https://github.com/tatsu-lab/stanford_alpaca}}.
\newblock


\bibitem[Team et~al\mbox{.}(2024)]%
        {team2024qwen2}
\bibfield{author}{\bibinfo{person}{Qwen Team} {et~al\mbox{.}}} \bibinfo{year}{2024}\natexlab{}.
\newblock \showarticletitle{Qwen2 technical report}.
\newblock \bibinfo{journal}{\emph{arXiv preprint arXiv:2407.10671}} (\bibinfo{year}{2024}).
\newblock


\bibitem[Touvron et~al\mbox{.}(2023)]%
        {touvron2023llamaopenefficientfoundation}
\bibfield{author}{\bibinfo{person}{Hugo Touvron}, \bibinfo{person}{Thibaut Lavril}, \bibinfo{person}{Gautier Izacard}, \bibinfo{person}{Xavier Martinet}, \bibinfo{person}{Marie-Anne Lachaux}, \bibinfo{person}{Timothée Lacroix}, \bibinfo{person}{Baptiste Rozière}, \bibinfo{person}{Naman Goyal}, \bibinfo{person}{Eric Hambro}, \bibinfo{person}{Faisal Azhar}, \bibinfo{person}{Aurelien Rodriguez}, \bibinfo{person}{Armand Joulin}, \bibinfo{person}{Edouard Grave}, {and} \bibinfo{person}{Guillaume Lample}.} \bibinfo{year}{2023}\natexlab{}.
\newblock \bibinfo{title}{LLaMA: Open and Efficient Foundation Language Models}.
\newblock
\showeprint[arxiv]{2302.13971}~[cs.CL]
\urldef\tempurl%
\url{https://arxiv.org/abs/2302.13971}
\showURL{%
\tempurl}


\bibitem[Unell et~al\mbox{.}(2025)]%
        {unell2025real}
\bibfield{author}{\bibinfo{person}{Alyssa Unell}, \bibinfo{person}{Mehr Kashyap}, \bibinfo{person}{Michael Pfeffer}, {and} \bibinfo{person}{Nigam Shah}.} \bibinfo{year}{2025}\natexlab{}.
\newblock \showarticletitle{Real-World Usage Patterns of Large Language Models in Healthcare}.
\newblock \bibinfo{journal}{\emph{medRxiv}} (\bibinfo{year}{2025}), \bibinfo{pages}{2025--05}.
\newblock


\bibitem[Wu et~al\mbox{.}(2023)]%
        {fastserve}
\bibfield{author}{\bibinfo{person}{Bingyang Wu}, \bibinfo{person}{Yinmin Zhong}, \bibinfo{person}{Zili Zhang}, \bibinfo{person}{Shengyu Liu}, \bibinfo{person}{Fangyue Liu}, \bibinfo{person}{Yuanhang Sun}, \bibinfo{person}{Gang Huang}, \bibinfo{person}{Xuanzhe Liu}, {and} \bibinfo{person}{Xin Jin}.} \bibinfo{year}{2023}\natexlab{}.
\newblock \showarticletitle{Fast distributed inference serving for large language models}.
\newblock \bibinfo{journal}{\emph{arXiv preprint arXiv:2305.05920}} (\bibinfo{year}{2023}).
\newblock


\bibitem[Wu et~al\mbox{.}(2024)]%
        {wu2024dlora}
\bibfield{author}{\bibinfo{person}{Bingyang Wu}, \bibinfo{person}{Ruidong Zhu}, \bibinfo{person}{Zili Zhang}, \bibinfo{person}{Peng Sun}, \bibinfo{person}{Xuanzhe Liu}, {and} \bibinfo{person}{Xin Jin}.} \bibinfo{year}{2024}\natexlab{}.
\newblock \showarticletitle{$\{$dLoRA$\}$: Dynamically orchestrating requests and adapters for $\{$LoRA$\}$$\{$LLM$\}$ serving}. In \bibinfo{booktitle}{\emph{18th USENIX Symposium on Operating Systems Design and Implementation (OSDI 24)}}. \bibinfo{pages}{911--927}.
\newblock


\bibitem[Ye et~al\mbox{.}(2025)]%
        {ye2025flashinfer}
\bibfield{author}{\bibinfo{person}{Zihao Ye}, \bibinfo{person}{Lequn Chen}, \bibinfo{person}{Ruihang Lai}, \bibinfo{person}{Wuwei Lin}, \bibinfo{person}{Yineng Zhang}, \bibinfo{person}{Stephanie Wang}, \bibinfo{person}{Tianqi Chen}, \bibinfo{person}{Baris Kasikci}, \bibinfo{person}{Vinod Grover}, \bibinfo{person}{Arvind Krishnamurthy}, {et~al\mbox{.}}} \bibinfo{year}{2025}\natexlab{}.
\newblock \showarticletitle{Flashinfer: Efficient and customizable attention engine for llm inference serving}.
\newblock \bibinfo{journal}{\emph{arXiv preprint arXiv:2501.01005}} (\bibinfo{year}{2025}).
\newblock


\bibitem[Zhang et~al\mbox{.}(2025)]%
        {elora}
\bibfield{author}{\bibinfo{person}{Hang Zhang}, \bibinfo{person}{Jiuchen Shi}, \bibinfo{person}{Yixiao Wang}, \bibinfo{person}{Quan Chen}, \bibinfo{person}{Yizhou Shan}, {and} \bibinfo{person}{Minyi Guo}.} \bibinfo{year}{2025}\natexlab{}.
\newblock \showarticletitle{Improving the Serving Performance of Multi-LoRA Large Language Models via Efficient LoRA and KV Cache Management}.
\newblock \bibinfo{journal}{\emph{arXiv preprint arXiv:2505.03756}} (\bibinfo{year}{2025}).
\newblock


\bibitem[Zhang et~al\mbox{.}(2023)]%
        {alpacare}
\bibfield{author}{\bibinfo{person}{Xinlu Zhang}, \bibinfo{person}{Chenxin Tian}, \bibinfo{person}{Xianjun Yang}, \bibinfo{person}{Lichang Chen}, \bibinfo{person}{Zekun Li}, {and} \bibinfo{person}{Linda~Ruth Petzold}.} \bibinfo{year}{2023}\natexlab{}.
\newblock \bibinfo{title}{AlpaCare: Instruction-tuned Large Language Models for Medical Application}.
\newblock
\showeprint[arxiv]{2310.14558}~[cs.CL]


\bibitem[Zheng et~al\mbox{.}(2023)]%
        {zheng2023efficiently}
\bibfield{author}{\bibinfo{person}{Lianmin Zheng}, \bibinfo{person}{Liangsheng Yin}, \bibinfo{person}{Zhiqiang Xie}, \bibinfo{person}{Jeff Huang}, \bibinfo{person}{Chuyue Sun}, \bibinfo{person}{Cody\_Hao Yu}, \bibinfo{person}{Shiyi Cao}, \bibinfo{person}{Christos Kozyrakis}, \bibinfo{person}{Ion Stoica}, \bibinfo{person}{Joseph~E Gonzalez}, {et~al\mbox{.}}} \bibinfo{year}{2023}\natexlab{}.
\newblock \showarticletitle{Efficiently Programming Large Language Models using SGLang.}
\newblock \bibinfo{journal}{\emph{arXiv preprint arXiv:2312.07104}} (\bibinfo{year}{2023}).
\newblock


\bibitem[Zhong et~al\mbox{.}(2024)]%
        {zhong2024distserve}
\bibfield{author}{\bibinfo{person}{Yinmin Zhong}, \bibinfo{person}{Shengyu Liu}, \bibinfo{person}{Junda Chen}, \bibinfo{person}{Jianbo Hu}, \bibinfo{person}{Yibo Zhu}, \bibinfo{person}{Xuanzhe Liu}, \bibinfo{person}{Xin Jin}, {and} \bibinfo{person}{Hao Zhang}.} \bibinfo{year}{2024}\natexlab{}.
\newblock \showarticletitle{$\{$DistServe$\}$: Disaggregating prefill and decoding for goodput-optimized large language model serving}. In \bibinfo{booktitle}{\emph{18th USENIX Symposium on Operating Systems Design and Implementation (OSDI 24)}}. \bibinfo{pages}{193--210}.
\newblock


\end{thebibliography}
